\documentclass[11pt,letterpaper]{article}
\usepackage[margin=1in]{geometry}
\usepackage[T1]{fontenc}
\usepackage{iftex}
\ifPDFTeX
  \usepackage[utf8]{inputenc}
\fi
\usepackage{lmodern}
\usepackage{amsmath,amssymb,amsthm,mathtools,bm}
\usepackage{booktabs,tabularx,array,multirow}
\usepackage{graphicx}
\usepackage{flafter,placeins}

\usepackage[dvipsnames]{xcolor}
\usepackage{tikz}
\usetikzlibrary{arrows.meta,calc,positioning,fit,shapes.geometric}
\usepackage{enumitem}
\usepackage[authoryear,round]{natbib}
\usepackage{xurl}
\usepackage{hyperref}
\usepackage[nameinlink,noabbrev]{cleveref}
\graphicspath{{figures/}}
\definecolor{CGRBlue}{HTML}{2F5D8C}
\definecolor{CGROrange}{HTML}{D58A3A}
\definecolor{CGRInk}{HTML}{27313A}
\hypersetup{
  pdftitle={Coarse-Graining Recursive Expert Computation: Continuation-Relative Interfaces and Failure Modes},
  pdfauthor={Shihua Yu; Yanzhi Wang; Tianchi Chen},
  pdfsubject={Continuation-relative evaluation of computational replacement},
  pdfkeywords={conditional computation, coarse-graining, continuation},
  hidelinks
}

\newcommand{\R}{\mathbb{R}}
\newcommand{\simplex}{\Delta^{E-1}}
\newcommand{\teacher}{\Psi}
\newcommand{\student}{\Phi_{\theta}}

\newcommand{\TV}{\operatorname{TV}}
\newcommand{\E}{\mathbb{E}}
\newcommand{\norm}[1]{\left\lVert #1\right\rVert}

\newcolumntype{Y}{>{\raggedright\arraybackslash}X}

\title{Coarse-Graining Recursive Expert Computation:\\Continuation-Relative Interfaces and Failure Modes}
\author{%
  Shihua Yu\textsuperscript{1}\qquad
  Yanzhi Wang\textsuperscript{1}\qquad
  Tianchi Chen\textsuperscript{2,*}}
\date{}

\begin{document}
\maketitle
\begingroup
\renewcommand{\thefootnote}{}
\makeatletter
\renewcommand{\@makefntext}[1]{\noindent#1}
\let\@footnotetext\H@@footnotetext
\makeatother
\footnotetext{%
  \textsuperscript{1}Department of Electrical and Computer Engineering, Northeastern University, Boston, MA, USA.\par
  \textsuperscript{2}Merck \& Co., Inc., Rahway, NJ, USA.\par
  Shihua Yu and Tianchi Chen are co-first authors.\par
  \textsuperscript{*}Corresponding author: \href{mailto:tianchi.chen@merck.com}{tianchi.chen@merck.com}.}
\endgroup
\vspace{-1.5em}
\begin{abstract}
Computational coarse-graining aims to replace a sequence of updates with a cheaper module while preserving task performance. We present a continuation-relative evaluation framework that assesses replacements through the computation that follows them, using downstream task quality and total execution cost as joint criteria. The framework distinguishes boundary representations, summaries of skipped computation, executed operations, and downstream loss. It makes interfaces and permitted deployment information explicit, combines fine-path references with directly trained controls using the same shortened architecture, and accounts for selection overhead. An exact loss decomposition separates replacement effects at a fixed downstream action from effects of changing that action. A finite construction shows how retained information can reduce the stipulated cost of zero-error prediction under fixed information and controller restrictions. Experiments in two frozen pretrained language models show that representation fidelity and task loss can favor different fitting objectives. In a prespecified comparison on a 1.7-billion-parameter model, next-token-loss training lowers in-distribution negative log-likelihood by 0.1901 nats per scored token compared with reconstructing the skipped blocks' output, using the same optimizer and update count. This ranking reverses in all nine paired comparisons under distribution shift. In a recursive expert prototype, the jointly adapted candidate lowers task error relative to frozen-bank and router-only baselines, but trails directly trained controls, and the tested input-only selectors fail the joint quality-cost criteria. Together, the protocol and studies provide explicit criteria and diagnostics for distinguishing useful computational replacement from local approximation gains within specified interfaces and evaluation settings.
\end{abstract}

\section{Introduction}\label{fmt:introduction}
Recursive networks, iterative solvers, and denoising models predict through repeated state updates. Conditional computation also uses the current state to choose operations. Replacing several updates with a cheaper module changes downstream information and may change which operations run next. Closely matching the original module's output therefore need not improve task performance or cost.

We ask: \emph{how should a computational replacement be evaluated in terms of downstream task quality and total cost?} We call the original updates the \emph{fine path}, their substitute a \emph{macro}, and all computation after the replacement boundary the \emph{continuation}. We develop an evaluation framework that makes this continuation explicit, derive conditional results, and test the framework in controlled studies:
\begin{itemize}[leftmargin=*,itemsep=3pt]
\item An evaluation protocol specifies continuation inputs, two fine-path quality references, controls trained with the same reduced computation graph, and selection costs using permitted deployment inputs. These comparisons measure substitution effects, the alternative of further fine-path training, benefits of the training procedure, and selection overhead.
\item Conditional analysis distinguishes the state at the replacement boundary, statistics of the skipped updates, executed operations, and task loss. An exact loss-gap decomposition separates the effect of replacement at a fixed continuation action from the effect of changing that action. A finite construction illustrates how retained information affects attainable task quality and assigned cost.
\item Controlled comparisons distinguish fitting-objective rankings from overall quality and cost. In two frozen pretrained language models, objective rankings depend on the evaluation distribution. In a recursive-expert prototype, the candidate loses to stronger training controls and neither tested input-only selector achieves the required quality and cost improvements.
\end{itemize}
The construction, learned models, and deployment measurements answer complementary questions about retained information, task performance, and whole-system quality and cost.
\subsection{Relation to prior work}\label{sec:related}
The framework builds on work on repeated computation, module replacement, and fidelity defined by downstream decisions. We combine these ideas in an explicit evaluation protocol and study their consequences in controlled experiments.

\paragraph{Repeated and conditional computation.}
Flow matching and rectified flow learn transport dynamics \citep{lipman2023flow,liu2022rectified}; distillation, consistency, shortcut, and mean-flow methods reduce repeated evaluations \citep{salimans2022progressive,song2023consistency,frans2025shortcut,geng2025mean}. Recursive and self-conditioned flow maps pursue related compression \citep{huang2026recursive,lee2026fmlm,yoo2026fixedpoint}. Looped Flows couples recurrent reasoning with local denoising, whereas LiFT supervises a shared core along computational depth within each flow evaluation \citep{suleymanzade2026loopedflows,derakhshani2026lift}. Universal Transformers, deep equilibrium models, consistency DEQs, and LoopFormer reuse parameters across depth \citep{dehghani2019universal,bai2019deq,lin2026cdeq,jeddi2026loopformer}. Mixture-of-Recursions routes tokens to different numbers of applications of a shared stack \citep{bae2025mor}. Sparse mixture-of-experts (MoE) routing \citep{shazeer2017moe,fedus2022switch} combines with recurrence in LoopMoE and The Thinking Pixel \citep{chen2026loopmoe,sun2026thinking}. We evaluate replacements in light of these precedents; recurrence, conditional selection, and flow compression are established ingredients.

\paragraph{Replacement and training controls.}
Latent Flow Transformer and LLM-Streamline learn block substitutes \citep{wu2025latentflow,chen2024streamline}. ReplaceMe primarily transforms a surviving block's multilayer perceptron (MLP) output and also studies full-block alignment; Ghosted Layers uses a linear operator on the residual stream, closer to the substitute tested here \citep{shopkhoev2025replaceme,yun2026ghosted}. BERT-of-Theseus trains replacement modules with task loss through frozen components of the original network \citep{xu2020theseus}. Expert pruning, Mixture-of-Depths, and CALM address reconstruction, skipping, and calibrated exits \citep{lu2024notallexperts,raposo2024mod,schuster2022calm}. Retraining the reduced computation graph is an established strong control \citep{liu2019rethinking,xu2026smallllms}; teacher misspecification can limit imitation \citep{hara2026misspecification}. Student-friendly teachers, student-aware routers, System-1.5, and recursive shortcut reasoning adapt or repair the distilled computation \citep{park2021sftn,kim2025everyexpert,wang2025system15,shankar2026shortcut}. Routing signatures summarize activations without necessarily retaining execution order \citep{avinash2026routing,li2025expertsignatures}; DYNAMOD jointly learns prediction and gating under a cost budget \citep{nan2017dynamic}. In our experiments, the selector that chooses between complete paths is fitted after those paths are frozen.

\paragraph{Downstream usefulness.}
Value equivalence, value-aware modeling, and counterfactual quotient models assess fidelity relative to specified decisions \citep{grimm2020value,farahmand2017vaml,chen2026cqm}; empirical work on objective mismatch studies a related distinction \citep{lambert2020objective}. Model stitching already compares affine activation matching with task-loss fitting through a fixed task map \citep{csiszarik2021similarity}. Distillation studies distinguish agreement with a teacher from task performance \citep{stanton2021does}; end-to-end sparse dictionary learning compares activation reconstruction with preserving downstream output distributions \citep{braun2024e2e}. We apply these distinctions to recursive replacement, using two fine-path references, controls with the same computation graph, separate measurements of predicted paths and executed operations, and costs that include selection from permitted deployment inputs. Our experiments compare fitting objectives within specified graphs; they do not reproduce complete implementations of these prior methods.
\FloatBarrier
\section{Continuation-Relative Evaluation Framework}\label{fmt:contract}
A replacement is evaluated at a specified computational boundary, through a fixed continuation and on a specified task distribution. Fine and macro paths may produce different representations, but must supply compatible inputs to the same continuation, including any required history.

\subsection{Evaluation protocol}
Four rules make the comparison interpretable:
\begin{enumerate}[label=(\alph*),leftmargin=*,itemsep=3pt]
\item \textbf{Shared interface.} Specify boundary inputs, fix continuation parameters, and define permitted downstream choices, called the \emph{action menu}. Actions must have the same meaning for both paths, directly or through a fixed alignment learned on separate calibration data, without rematching individual test examples.
\item \textbf{Same-graph controls.} Train the same reduced computation graph from initialization under specified budgets, and identify any components kept frozen. Matching update counts or forward operations alone does not match total training cost.
\item \textbf{Quality references.} When both paths are adapted together, compare the macro with its own fine path and with a fine path trained further without a macro. These measure substitution effects and the alternative use of adaptation resources, respectively. Specify task loss, allowed degradation, evaluation distribution, and aggregation.
\item \textbf{Selection and deployment cost.} A selector may use only information permitted and available at deployment decision time. Include prediction, example partitioning, and dispatch costs. An offline oracle using evaluation labels estimates selection's potential under extra information, not deployment savings.
\end{enumerate}
Each application also specifies the retained state and history, assumptions about whether they determine future updates, the information permitted and available to each component, and the cost model. Changing the continuation or evaluation distribution requires reassessment. A continuation with no downstream choice has a single action, so the reallocation term below is zero. Tool side effects and asynchronous or hidden environmental state require an expanded interface. The protocol makes comparisons interpretable but does not guarantee success.
\subsection{State, path, execution, and continuation}\label{fmt:interfaces}\label{fmt:state-signature}
The \emph{boundary state} carries information into future computation; a \emph{path statistic} summarizes skipped updates; and \emph{execution} records operations actually performed. Usefulness is measured by task loss and complete cost. Predicting which experts the skipped updates would have used does not mean executing them.

For $E$ experts and $K$ fine steps, write $z_{i+1}=F_i(z_i,c,b_i)$, where $z_i$ is visible state, $c$ fixed context, and $b_i$ any needed history. \emph{Closure} means that these retained inputs, together with step $i$, contain all information needed to determine the next update. If histories with the same visible state produce different futures, a state-only predictor cannot represent both; the relevant history must be included and carried forward (\cref{prop:closure}).

An interval's \emph{occupancy} is the average weight assigned to each expert. With equally spaced steps, $Q_I=n^{-1}\sum_{i\in I}\pi_i$, where $n=|I|>0$ and $\pi_i$ is a nonnegative $E$-vector of expert weights summing to one. Adjacent occupancies combine by length-weighted averaging, evaluating the second interval from the state reached after the first (\cref{eq:cocycle}). Occupancy records amounts, not order: $A(z)=2z$ followed by $B(z)=z+1$ gives $2z+1$, whereas reversing them gives $2z+2$, despite identical occupancy. \Cref{fmt:appendix-start} defines an order-sensitive extension using adjacent routing pairs, accounts for assignments dropped by capacity limits, and gives composition laws. This extension need not recover a full routing sequence, and the prototype does not train it.

\begin{figure}[!htbp]
\centering
\resizebox{0.98\linewidth}{!}{\begin{tikzpicture}[x=1cm,y=1cm,font=\sffamily\fontsize{8.5}{10}\selectfont,
  box/.style={draw=CGRInk,rounded corners=2pt,align=center,minimum height=0.85cm,inner sep=4pt},
  flow/.style={-{Latex[length=2mm]},semithick,draw=CGRInk}]
\node[anchor=west,font=\sffamily\small\bfseries] at (0,2.0) {A\quad Different prefixes, shared continuation};
\node[box,text width=1.55cm] (input) at (0.95,0) {Input $x$\\Encoder $z_0$};
\node[box,draw=CGRBlue,text width=3.05cm] (fine) at (4.1,0.75) {Fine prefix\\Two expert updates};
\node[box,draw=CGROrange,text width=3.05cm] (macro) at (4.1,-0.75) {Macro replacement\\One learned map};
\coordinate (boundary) at (8.0,0);
\node[align=center,text width=1.7cm] at (6.65,1.38) {$z_F,q_F$\\Measured\\occupancy};
\node[align=center,text width=1.7cm] at (6.65,-1.4) {$z_M,q_M$\\Predicted\\occupancy};
\node[box,text width=1.7cm] (router) at (9.75,0.45) {Router\\reads $z,q$};
\node[box,text width=2.1cm] (expert) at (12.75,0.45) {Selected expert\\+ task head\\Reads state $z$};
\node[draw=CGRInk!45,rounded corners=2pt,fit=(router)(expert),inner xsep=7pt,inner ysep=8pt] (shared) {};
\node[anchor=south] at (shared.north) {Shared parameters};
\draw[flow] (input.east) -- ++(0.3,0) |- (fine.west);
\draw[flow] (input.east) -- ++(0.3,0) |- (macro.west);
\draw[flow,draw=CGRBlue] (fine.east) -- (7.65,0.75) -- (boundary);
\draw[flow,draw=CGROrange] (macro.east) -- (7.65,-0.75) -- (boundary);
\fill[CGRInk] (boundary) circle (1pt);
\draw[flow] (boundary) |- (router.west);
\draw[flow] (router.east) -- node[above] {$a$} (expert.west);
\draw[flow] (boundary) -- (8.0,-0.65) -- node[below,font=\footnotesize] {state $z$} (12.75,-0.65) -- (expert.south);
\node[font=\footnotesize,text=CGRInk!80,anchor=west] at (0,-2.2) {Alternative paths: selecting a macro does not execute the skipped fine updates.};
\node[anchor=west,font=\sffamily\small\bfseries] at (0,-2.65) {B\quad Evaluate the complete replacement};
\node[box,text width=6.25cm,minimum height=1.05cm] at (3.25,-3.55)
  {\textbf{Shared interface}\\Compatible inputs; same action meanings};
\node[box,text width=6.25cm,minimum height=1.05cm] at (10.55,-3.55)
  {\textbf{Task-quality references}\\Own fine path and continued fine-only model};
\node[box,text width=6.25cm,minimum height=1.05cm] at (3.25,-4.8)
  {\textbf{Same-graph DIRECT controls}\\Direct training under specified budgets};
\node[box,text width=6.25cm,minimum height=1.05cm] at (10.55,-4.8)
  {\textbf{Deployment utility}\\Permitted selector inputs; selection + path cost};
\end{tikzpicture} }
\caption{\textbf{Evaluate a replacement through its continuation and complete execution cost.} (A) The recursive prototype illustrates the shared interface. Boundary states are $z_F,z_M$; $q_F$ is the fine interval's measured occupancy $Q_I$, and $q_M$ is the macro's prediction without executing those updates. Both paths supply state and occupancy to the same router. Its action $a$ selects an expert, which receives the state and feeds its output to the task head; occupancy is used only by the router. An input-only path selector, distinct from this expert router, can choose between the frozen complete paths. (B) The evaluation protocol compares task quality with both fine references when paths are jointly adapted, includes direct training of the same reduced graph, and charges selector prediction, example partitioning, and dispatch along with path execution. The diagram specifies comparisons, not a guarantee of improvement.}
\label{fig:continuation-interface}
\end{figure}

\paragraph{A finite constructive case.}\label{fmt:finite-construction}
Retained information can change attainable risk and assigned cost under fixed information restrictions. For $k\in\{2,3,4\}$ blocks, independent fair bits $(A_i,B_i,P_i)$ determine a query: answer $A_i\oplus B_i$ (exclusive-or) if $P_i=0$, and $B_i$ otherwise. A macro observes $(A_i,B_i)$, not $P_i$, and writes one bit per block. For each summary, a controller reused across blocks reads only that bit and $P_i$. For each of these four possible input pairs, it returns 0, returns 1, or calls an expert for the answer, giving $3^4=81$ deterministic controllers. The assigned fee is $1.60$ per input plus $1.10$ per expert call; risk averages query errors over all inputs. The XOR summary answers $P_i=0$ queries directly and needs expert calls on only half the queries on average; $A_i$ gives no information about either answer. Enumerating all inputs and controllers, zero risk therefore costs $1.60+1.10k/2$ on average with the XOR summary and $1.60+1.10k$ with the equal-width $A_i$ summary. With expected fee capped at the XOR summary's minimum zero-risk fee, the $A_i$ summary's minimum attainable risk is $0.25$. A block-index-dependent controller with the XOR summary ties the minimum zero-risk fee, but not necessarily the full risk--fee frontier. These restricted controllers and stipulated fees do not demonstrate learned compression or hardware savings (\cref{app:scale-k-r-sealed}).
\subsection{Conditional analysis of the continuation gap}\label{fmt:limits}\label{fmt:six-conditions}\label{fmt:continuation}
To separate replacement from changes in downstream action, fix an input--target pair $(x,y)$ and freeze both prefixes, the shared continuation, router, readout, and deterministic tie rule. Let $\mathcal A$ be a common finite nonempty action menu, $o_j(a)$ the output after prefix $j\in\{F,M\}$ and action $a$, and $a_j$ the router's chosen action; $F$ denotes the fine prefix and $M$ the macro. The router uses the inputs produced by each prefix, without access to $y$ or action-menu outputs computed for offline diagnostics. For squared loss, define
\begin{equation}
\ell_j(a)=\tfrac12\|o_j(a)-y\|_2^2,\qquad
m_j=\min_{a\in\mathcal A}\ell_j(a),\qquad
\mathfrak r_j=\ell_j(a_j)-m_j\ge0.
\label{fmt:continuation-loss}
\end{equation}
The minimum $m_j$ is computed offline using the target. Router \emph{regret} $\mathfrak r_j$ is the chosen action's excess loss above this minimum for the current input. Adding and subtracting the macro loss at the fine path's action gives
\begin{align}
\ell_M(a_M)-\ell_F(a_F)
&=\underbrace{\ell_M(a_F)-\ell_F(a_F)}_{\text{fixed-action term}}
 +\underbrace{\ell_M(a_M)-\ell_M(a_F)}_{\text{reallocation term}}\nonumber\\
&=m_M-m_F+\mathfrak r_M-\mathfrak r_F.
\label{fmt:continuation-decomposition}
\end{align}
The fixed-action term measures replacement's effect on loss with the fine action retained. Reallocation measures the further loss change when the macro's router chooses its own action, which may compensate for or worsen replacement error. This exact frozen-model identity does not identify training's causal effects.

Uniform agreement across actions bounds the difference between minimum losses, but not router regret. If $\max_a|\ell_M(a)-\ell_F(a)|\le\varepsilon_\ell$, then $|m_M-m_F|\le\varepsilon_\ell$. For Euclidean squared loss, suppose that for every action the macro and fine outputs differ in norm by at most $\varepsilon_o$, and the fine output differs from the target in norm by at most $B_{\rm res}$. Then one may take $\varepsilon_\ell=B_{\rm res}\varepsilon_o+\varepsilon_o^2/2$: the first term scales the output mismatch by the fine path's existing error, and the second accounts for the mismatch's squared size. Minibatch mean discrepancies do not establish these uniform bounds or control $\mathfrak r_M-\mathfrak r_F$. Exact equality of \emph{all} specified inputs to the same deterministic continuation makes both gap terms zero; matching selected statistics alone need not do so (\cref{prop:continuation-standard}).

The remaining conditional results identify further limits (\cref{app:signature-analysis}). Bounds on error relative to the fine path require each shorter replacement to be accurate where it is evaluated, including at states predicted by earlier replacements. Endpoint agreement need not identify routing weights, even when all routing weights are positive. The selected top-$k$ experts remain unchanged if every routing logit changes by less than half the positive gap between the $k$-th and $(k+1)$-th largest logits, before capacity limits are applied. The small set of experts actually executed differs from an occupancy prediction that may assign positive weight to every expert. The supplementary information (SI) states each result's assumptions and interpretation. Physical time, generative-flow time, and recursive depth are distinct coordinates (\cref{sec:three-clocks}); the experiments below do not implement generative flow matching.
\section{Study Design}\label{fmt:prototype}
The finite construction shows how retained information can reduce assigned cost at fixed risk. The learned studies ask complementary questions: language-model substitutions compare fitting objectives within frozen networks, while the recursive-expert prototype also tests routing and whole-system quality and cost. These experiments do not reproduce complete external compression methods for a head-to-head comparison.

\subsection{Recursive prototype and controls}
The prototype predicts two real-valued targets from four inputs. In the original synthetic task, an input-dependent condition selects between two nonlinear target formulas (\cref{sec:unified-v3}). An encoder maps the inputs to a width-16 state. A fine prefix executes two updates, each selecting one of eight shared residual experts (top-1 routing); a width-24 macro predicts a replacement state and the skipped prefix's occupancy. Both paths feed the same router, which chooses one continuation expert before a two-output readout. The router receives each path's state and occupancy, whereas the selected expert receives only the state. Reported predictions use \emph{hard dispatch}: each routing decision executes only the selected expert. Training uses a straight-through estimator for prefix routing gradients and probability-weighted individual continuation losses, not the exact gradient of hard dispatch. Architecture, coefficients, gradient paths, and load balancing are specified in \cref{app:synthetic-model}.

The base objective combines fine- and macro-path task mean squared error (MSE), boundary-state matching, and load balancing. Additional terms match the fine path's occupancy (\emph{occupancy anchoring}), make a macro agree with a composition of shorter macros (\emph{self-composition}), or match continuation outputs for corresponding actions (\emph{response matching}). Relevance-weighted response matching gives greater weight to actions with lower fine-path task loss; these label-informed training weights are held fixed during gradient computation. \Cref{tab:arms} separates these objectives from restrictions on trainable components and training the same reduced graph from initialization. Only the macro replacing both fine updates is directly supervised by the fine path; maps for single-update spans serve as auxiliary factors in the composition objective and are not calibrated to individual fine steps.

\begin{table}[!htbp]
\centering\small
\caption{\textbf{Controls distinguish objectives, trainable components, and budgets.} The base objective combines task loss, boundary-state matching, and load balancing. DIRECT trains the reduced graph from initialization; fine-only paths provide quality references.}
\label{tab:arms}
\begin{tabularx}{\linewidth}{@{}>{\raggedright\arraybackslash}p{0.22\linewidth}>{\raggedright\arraybackslash}p{0.17\linewidth}Y@{}}
\toprule
Method & Trainable parts & Objective and comparison role \\
\midrule
JOINT-END & all & Base objective; boundary-state-supervised baseline. \\
SIG & all & Base plus occupancy anchoring and state/occupancy composition. \\
RESP-ONLY & all & Base plus response matching, weighted uniformly over actions. \\
SHORTCUT-CONT & all & Base plus $0.1$ state-composition MSE and $0.1$ MSE between each path's selected output. \\
RELEVANT & all & Base plus anchoring, composition, and relevance-weighted response matching; development-selected candidate. \\
ROUTER-RELEVANT & macro and routers & Full RELEVANT objective; restricted router adaptation. \\
FROZEN-RELEVANT & macro only & RELEVANT penalties; no fine task loss; macro-only load term. \\
DIRECT-DATA / DIRECT-COMPUTE & reduced graph from initialization & Task loss and load balancing; update-matched / affine-operation-matched controls. \\
Continued fine-only / pretrained fine & fine path & Task loss and load balancing; continued-training / starting-checkpoint references. \\
\bottomrule
\end{tabularx}
\end{table}
\FloatBarrier

\subsection{Budgets, references, and decision rules}\label{fmt:synthetic-evaluation}\label{fmt:pass-rules}
A \emph{parent} is an independently initialized and pretrained fine model. Two development parents are used to select the candidate before evaluation on five new confirmation parents. Methods initialized from a given parent share its checkpoint and adaptation minibatches. Five additional synthetic target families follow protocols specified after the original results but fixed before their own runs. Uncertainty in the synthetic studies is assessed across parents, not examples or timing repeats. Language-model seeds instead vary fitting order within one frozen model.

Fine pretraining and adaptation each use 1,200 Adam updates. DIRECT-DATA trains the identical reduced graph from initialization for 2,400 updates. DIRECT-COMPUTE receives 7,910 updates, approximately matching the candidate's total forward-pass affine multiply--accumulate (MAC) operations during pretraining and adaptation. DIRECT uses uninterrupted Adam; methods initialized from pretrained parents reset it between phases. These budgets do not match total training cost. At deployment, macro and fine paths use 2,840 and 5,176 affine MACs per example, respectively, excluding non-affine operations and dispatch. Data, seeds, evaluated populations, and budget limitations are detailed in \cref{sec:unified-v3}.

For path $b$, parent $p$, split $s$, and two-output example $i$, let $\ell_{b,p,s,i}=\|\widehat y_{b,p,s,i}-y_{p,s,i}\|_2^2/2$. Task MSE averages examples and then parents equally:
\begin{equation}
L_{b,p,s}=\frac1{N_{p,s}}\sum_i\ell_{b,p,s,i},\qquad
L_{b,s}=\frac15\sum_{p=1}^5L_{b,p,s}.
\label{eq:contract-aggregation}
\end{equation}
The prespecified quality rule requires $L_{M,s}\le1.02\min(L_{\mathrm{own},s},L_{\mathrm{cont},s})$, comparing the macro with both its own jointly adapted fine path and the continued fine-only reference. The rule uses aggregate losses rather than mean parent-level ratios. Independently and identically distributed (IID) test data provide the primary evaluation; a shift in the input range is evaluated separately. Fixed-path IID utility additionally requires end-to-end median latency at batch size one no greater than $90\%$ of the macro's own fine-path latency, comparing means of per-parent medians. These thresholds are study-specific; latency under distribution shift was not measured.

\paragraph{Comparison and analysis scope.}\label{fmt:method-screen}\label{fmt:prototype-status}
The stronger criterion for improvement over alternative training methods requires at least $5\%$ lower mean IID MSE than every baseline named in \cref{sec:unified-v3}. This is a practical rule for the evaluated parents, not a powered statistical superiority test. The formal results also have limited empirical scope here. The tested prefix is memoryless, so the omitted-history bound adds no constraint. Predicted occupancy differs from executed occupancy; results about positive routing weights and sparse execution therefore do not certify diagnostics of the actual hard-dispatch path. Average composition penalties do not establish uniform bounds. The language-model studies evaluate next-token loss with a single continuation action and do not test the synthetic squared-loss routing bound.
\FloatBarrier
\section{Does Local Fidelity Predict Task Quality?}\label{fmt:realmodel}

Two frozen dense language models test whether a substitute that closely matches the original boundary state also improves downstream task performance, and whether the ranking persists under distribution shift. GPT-2 small and the Qwen3 replication use the same substitute class but separate protocols and statistical analyses. All comparisons are within model; raw negative log-likelihood (NLL) is not compared across tokenizers.

\paragraph{Shared replacement and controls.}
At boundary $(l,k)$, with zero-based block index $l$ and replacement length $k$, a residual affine map applied independently at each token position,
\[
g_{W,b}(h_l)=h_l+Wh_l+b,\qquad W\in\mathbb R^{d\times d},\quad b\in\mathbb R^d,
\]
replaces blocks $[l,l+k)$. Here $h_l$ is a token's hidden state before block $l$, $d$ is the model width, and $W,b$ are trainable substitute parameters. The remaining blocks, final normalization, and output head form the unchanged continuation. All original weights are frozen, but gradients through the continuation can train the substitute. The map acts on the full boundary state and is related to residual alignment; it does not reproduce a complete compression pipeline \citep{shopkhoev2025replaceme,yun2026ghosted}.

Six methods distinguish objectives and initialization: \textbf{DROP} uses the identity map; \textbf{FID-LS} fits boundary states by ridge least squares; \textbf{FID-SGD} uses Adam to minimize boundary-state MSE; \textbf{CONT-ZERO} minimizes next-token NLL from zero substitute parameters; \textbf{CONT-LS} starts the same task-loss fit from FID-LS; and \textbf{END-KL} minimizes the Kullback--Leibler divergence $\mathrm{KL}(p_{\mathrm{full}}\|p_{\mathrm{replaced}})$ between output distributions. With $W=b=0$, CONT-ZERO starts from the same identity map as DROP and trains only the substitute. Ridge optimizes its regularized fitting-data objective, which need not minimize held-out boundary error. Matching Adam updates and data does not match backward-pass cost; CONT-LS also requires ridge initialization.

MSE across boundary-state coordinates and output KL measure fidelity; NLL in nats per scored token measures task quality. Lower is better for all three. WikiText-2 provides fitting and IID evaluation text \citep{merity2016wikitext}; LAMBADA evaluates distribution shift \citep{paperno2016lambada}. FULL is the unadapted original model; no equally adapted full-model control is available. \Cref{app:realmodel,app:qwen3} specify each model's sampling, scoring, and uncertainty and report all six methods, including the stronger controls.

\subsection{GPT-2: the ordering depends on the evaluation text}
GPT-2 small has 12 blocks and width 768 \citep{radford2019gpt2}. The tested boundaries are $(6,2)$, $(4,2)$, $(8,2)$, $(4,4)$, and $(6,4)$. Adam-trained methods receive 400 updates with matched numbers of sampled token positions. CONT-ZERO has two seeds for minibatch sampling except at $(4,4)$, where seed 0 was not run; FID-SGD has one. Learning-rate selection and a documented change to the computing environment limit the comparison, and historical input revisions were not retained (\cref{app:realmodel}).

\begin{figure}[!htbp]
\centering
\includegraphics[width=\linewidth]{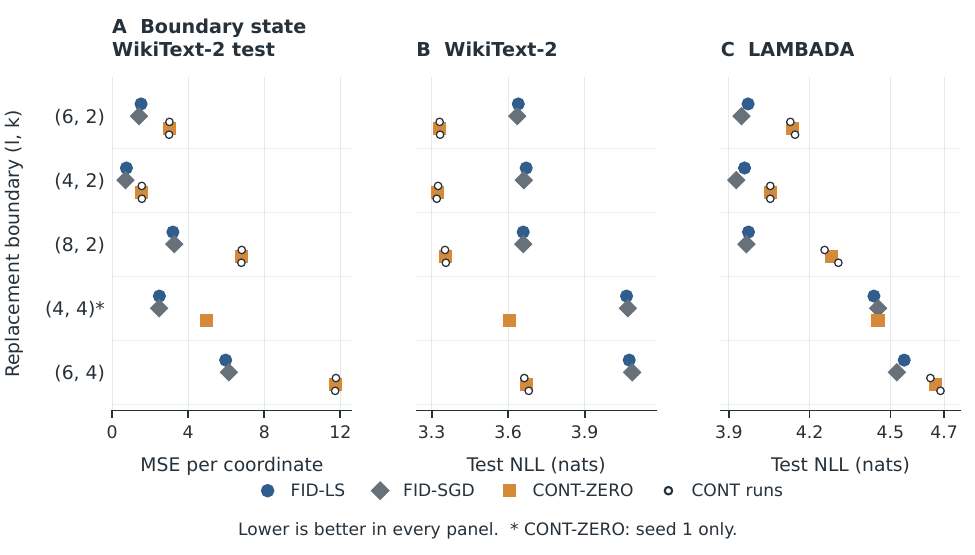}
\caption{\textbf{Boundary fidelity and task loss rank fitting objectives differently.} Rows are replacement boundaries within one GPT-2 checkpoint. Panels show (A) WikiText-2 test boundary MSE, (B) WikiText-2 test NLL, and (C) LAMBADA NLL; lower is better. Filled marks are means. FID-LS and FID-SGD have one run; CONT-ZERO has two except at $(4,4)$ (seed 1 only). Open circles show individual CONT-ZERO runs, not confidence intervals. The descriptive FID-SGD comparison matches Adam updates and sampled positions, not backward cost. \Cref{tab:realmodel} reports all six methods; the numerical supplement retains every metric.}
\label{fig:realmodel-objectives}
\end{figure}

CONT-ZERO has higher boundary MSE but lower IID NLL than FID-LS at every boundary; FID-LS has lower NLL on LAMBADA at every boundary (\cref{fig:realmodel-objectives}). The prespecified prediction that CONT-ZERO would trade higher boundary error for lower NLL on both distributions therefore fails, $0/5$; its IID-only criterion across all available seeds holds, $5/5$. Every available CONT-ZERO seed also beats FID-SGD on IID NLL despite larger MSE. This optimizer-matched comparison is descriptive, not a multi-seed significance test. FID-SGD has lower held-out boundary MSE than FID-LS at three boundaries, so ridge does not uniformly give the closest boundary match. CONT-ZERO improves on FULL's IID NLL at the three two-block boundaries, without establishing superiority across distributions.

Ridge remains a useful initialization: CONT-LS ends at or below CONT-ZERO's IID NLL at every boundary, while boundary error rises during fitting. The prespecified prediction of a positive KL--NLL correlation fails, $0/5$; the remaining predictions and fitting trajectories are reported in \cref{app:realmodel} and the numerical supplement. Exploratory changes to the fitting text alter the IID--shift trade-off but fail the prespecified criterion of doing no worse than FULL. The web source could not be identified reliably, so the change cannot be attributed to FineWeb-Edu or corpus breadth (\cref{app:realmodel-lookup}).
\subsection{Qwen3: objective rankings under distribution shift}\label{fmt:qwen3}

The replication uses Qwen3-1.7B-Base~\citep{qwen3technicalreport} at boundaries $(7,2)$, $(14,2)$, and $(21,2)$. Each stochastic method receives 256 Adam updates over the same 256 training windows, with three paired seeds changing only their order. There is no validation tuning or checkpoint selection. Exact input versions are recorded for this study; the historical GPT-2 inputs remain incompletely documented. \Cref{app:qwen3} gives the separate replication protocol and limits on uncertainty estimates.

\begin{figure}[!htbp]
\centering
\includegraphics[width=\linewidth]{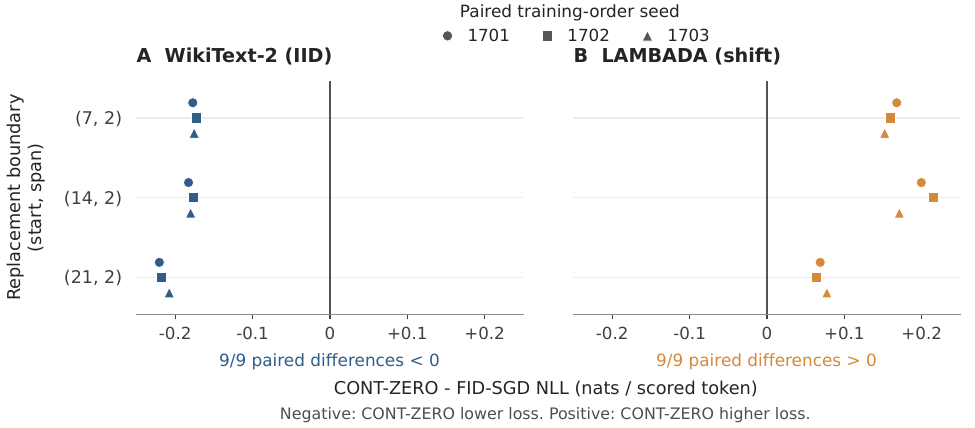}
\caption{\textbf{The Qwen3 objective ranking reverses under distribution shift.} Each panel reports CONT-ZERO minus FID-SGD NLL at the same three replacement boundaries and three paired training-order seeds. The common horizontal scale is in nats per scored token; negative values favor CONT-ZERO. Each mark is a paired run difference, giving nine marks per distribution from one pretrained model. Shapes identify seeds; vertical offsets separate marks. No confidence bars are shown. All IID differences are negative and all LAMBADA differences positive. NLL for all six methods and fidelity--task comparisons are in \cref{tab:qwen3-nll,fig:qwen3-fidelity-task}; every metric is preserved in the numerical supplement (\cref{app:numerical-companion}).}
\label{fig:qwen3-objectives}
\end{figure}

The prespecified primary comparison is held-out WikiText-2 NLL for CONT-ZERO minus FID-SGD, with optimizer and update count matched. CONT-ZERO lowers IID NLL in all nine boundary--seed pairs despite larger boundary MSE and output KL. Averaging boundaries equally within each seed and then averaging seeds gives $-0.1901$ nats per scored token, with a Student-$t$ $95\%$ interval of $[-0.1976,-0.1826]$ over the three seed means. The prespecified IID acceptance criteria are met. The separate paired-article interval is conditional on the fitted substitutes (\cref{app:qwen3}). All nine LAMBADA comparisons reverse direction: mean differences at the three boundaries are $+0.1600$, $+0.1953$, and $+0.0702$, so FID-SGD has lower NLL (\cref{fig:qwen3-objectives}). The CONT-ZERO advantage is therefore distribution-specific.

\begin{samepage}
The remaining controls limit this conclusion to the primary comparison. FID-LS has lower mean NLL than FID-SGD at every boundary on both distributions. CONT-LS lowers mean IID NLL relative to CONT-ZERO at all three boundaries, but raises mean LAMBADA NLL at $(21,2)$ ($3.7728$ versus $3.7181$). Both continuation-trained methods have higher mean LAMBADA NLL than untrained deletion at every boundary; every replacement method has higher NLL than the unadapted full model on that distribution. The continuation-trained methods improve on the full-model IID reference only at the first two boundaries. No equally adapted full-model control was run.
\par\end{samepage}

The study uses three training-order seeds, 21 fitting articles in source order, and one update budget. It neither replicates across independently pretrained models nor retests every GPT-2 prediction. Neither language-model study establishes transfer to MoE or flow matching, general superiority over the full model, or quality-preserving acceleration.
\FloatBarrier
\section{From Task Performance to Whole-System Utility}\label{fmt:evidence}
The recursive prototype tests whether a locally accurate substitute improves the complete computation. We compare complete paths, examine their loss gaps and auxiliary objectives, and include the cost of selecting a path from permitted deployment inputs.

\subsection{Complete-path comparisons and loss diagnostics}
\begin{figure}[!htbp]
 \centering
 \includegraphics[width=\linewidth]{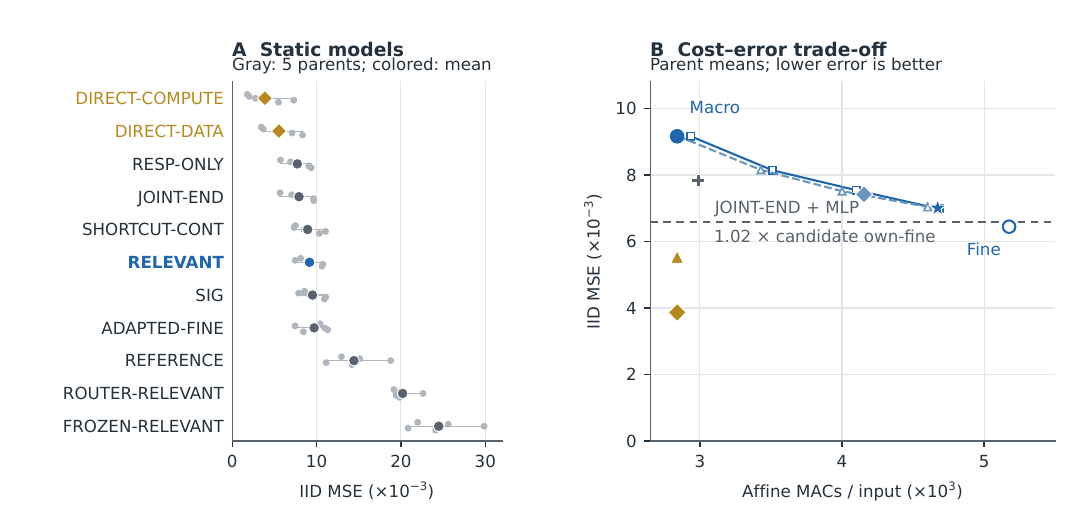}
 \par\smallskip
 \begingroup
\definecolor{ResultMLP}{HTML}{2166AC}
\definecolor{ResultLinear}{HTML}{6795BE}
\definecolor{ResultDirect}{HTML}{B68A21}
\small
\begin{tabular}{@{}l@{\quad}l@{\qquad}l@{\quad}l@{}}
\tikz[baseline=-0.5ex]{\draw[ResultMLP,line width=.7pt] (0,0)--(.45,0);\draw[ResultMLP,fill=white,line width=.7pt] (.20,-.04) rectangle (.28,.04);}
& MLP: DEV-frozen curve
& \tikz[baseline=-0.5ex]{\node[star,star points=5,star point ratio=2.5,fill=ResultMLP,draw=ResultMLP,minimum size=5pt,inner sep=0pt] {};}
& MLP: DEV-chosen point \\
\tikz[baseline=-0.5ex]{\draw[ResultLinear,dashed,line width=.7pt] (0,0)--(.45,0);\draw[ResultLinear,fill=white,line width=.7pt] (.19,-.04)--(.27,-.04)--(.23,.05)--cycle;}
& Linear: DEV-frozen curve
& \tikz[baseline=-0.5ex]{\draw[ResultLinear,fill=ResultLinear] (0,.06)--(.06,0)--(0,-.06)--(-.06,0)--cycle;}
& Linear: DEV-chosen point \\
\tikz[baseline=-0.5ex]{\draw[ResultDirect,fill=ResultDirect] (0,.07)--(.07,-.05)--(-.07,-.05)--cycle;}
& DIRECT-DATA
& \tikz[baseline=-0.5ex]{\draw[ResultDirect,fill=ResultDirect] (0,.06)--(.06,0)--(0,-.06)--(-.06,0)--cycle;}
& DIRECT-COMPUTE \\
\end{tabular}
\endgroup\space
 \caption{\textbf{Complete controls and the measured quality--work trade-off.}
 (A) IID MSE for each of five parents and their mean for each fixed path; lower is better.
 (B) Candidate fixed paths, selector operating points, and strong DIRECT
 controls, averaged over the same five parents. The horizontal limit is
 $1.02$ times the candidate's own-fine mean MSE: a necessary quality
 condition, not the complete quality--latency test. The plus mark is
 JOINT-END with its MLP selector at the development-chosen threshold.
 Selector thresholds are
 fixed on development data, without retuning quantiles on test data.
 Affine MACs exclude backward-pass work and non-affine deployment overhead,
 so the curves do not measure latency. No test-label oracle or population
 confidence band is shown. Label key:
 ADAPTED-FINE = continued fine-only; REFERENCE = pretrained fine reference;
 DEV = development split (DEV-chosen point: the threshold chosen on
 development data; DEV-frozen curve: thresholds fixed on development data).}
 \label{fig:unified-v3}
\end{figure}
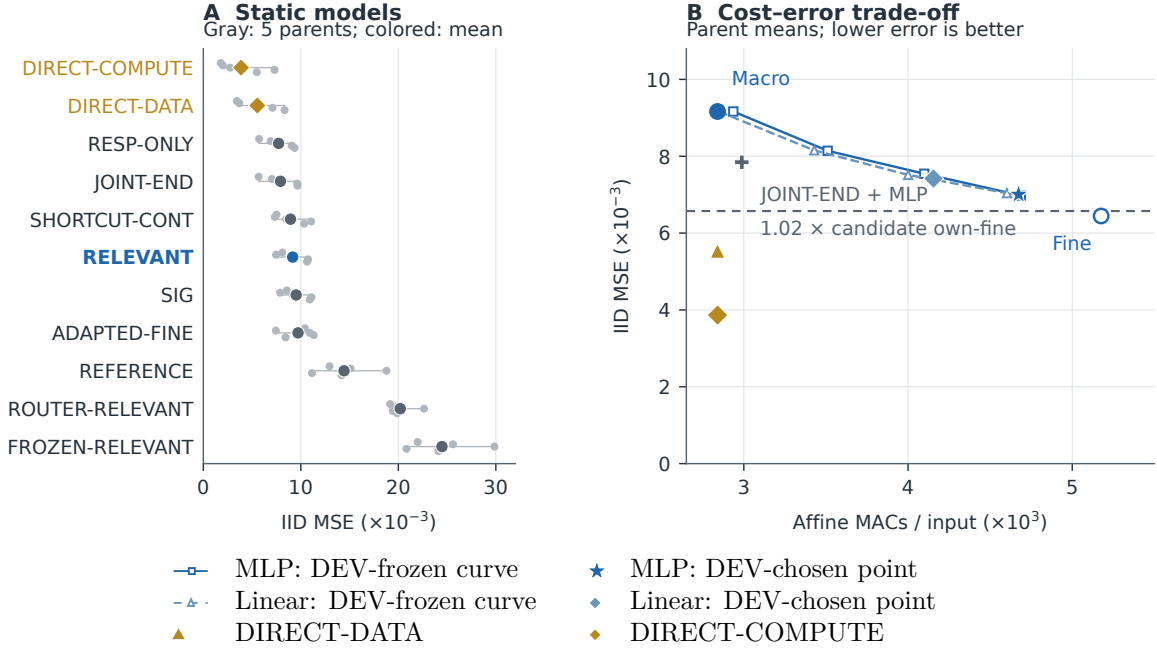

\Cref{fig:unified-v3} shows the original family's per-parent results and strong controls; \cref{tab:unified-fixed} reports all fixed-path losses on both splits. Jointly adapting more components improves on frozen-bank and router-only controls, but JOINT-END and both same-graph DIRECT controls have lower MSE than the candidate on every original-family parent and split. The candidate therefore fails the prespecified criterion for improvement over alternative methods. Across all six synthetic families, both DIRECT controls also have lower MSE on every parent--split pair. These families share input and output structure and do not represent six independent real domains.

\begin{figure}[p]
\centering
\includegraphics[width=\linewidth]{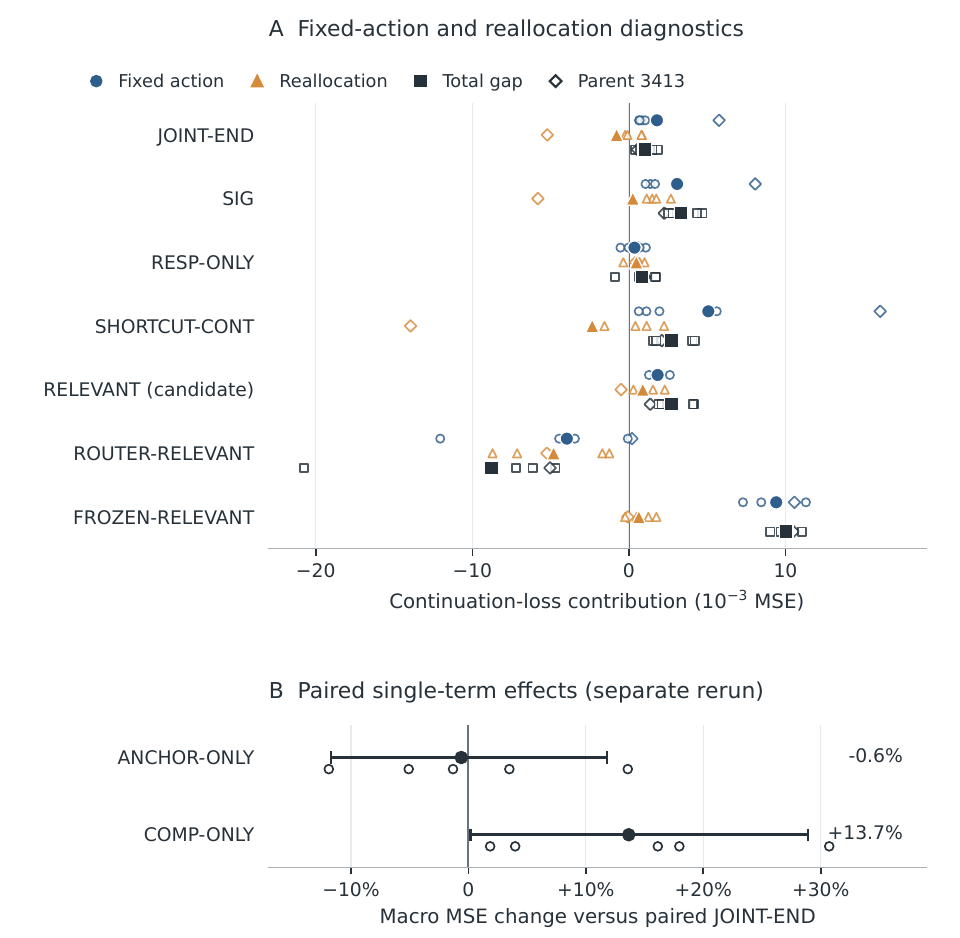}
\caption{\textbf{Loss-gap decomposition and auxiliary-objective effects in the original synthetic family.} (A) Original confirmation fits on IID data: fixed-action, reallocation, and total continuation-loss gaps in $10^{-3}$ MSE units. Positive components increase the macro-minus-fine loss gap. Large filled marks are arithmetic means; small open marks show the five parents. Diamonds identify parent 3413, whose negative reallocation determines the negative JOINT-END and SHORTCUT-CONT means. The decomposition describes frozen models and does not identify training effects. DIRECT has no corresponding fine path to decompose. (B) Separate exploratory single-term study with 1,200 fine-pretraining updates: percentage change in IID macro MSE relative to paired JOINT-END fits. Open marks show the five paired parent ratios; filled marks show geometric-mean changes. Bars are descriptive paired $95\%$ $t$-intervals computed on log ratios and back-transformed. These fits are not pooled with panel A. The occupancy-anchoring interval permits material effects in either direction. Full results and protocols are in \cref{app:decomp-shift,app:factor-decomp}.}
\label{fig:synthetic-diagnostics}
\end{figure}

The decomposition in \cref{fmt:continuation-decomposition} separates two contributions to the frozen model's loss gap (\cref{fig:synthetic-diagnostics}A). In the original family, the candidate's mean reallocation term is positive, so changing the continuation action raises its loss. Negative means for JOINT-END and SHORTCUT-CONT depend on one parent, while ROUTER-RELEVANT's negative gap accompanies higher fine-reference loss. A favorable gap term therefore need not imply good task performance. Macro-path regret is measured against its offline best action, not DIRECT. Reallocation can also reduce loss, and its direction changes across target families (\cref{app:decomp-shift,app:multifamily}).

\paragraph{Separating auxiliary fidelity from composition.}
The original comparison changes occupancy anchoring and state/occupancy composition together and, in some conditions, also changes response weighting. Its association with higher task loss cannot isolate the effect of fidelity matching. A separate study adds one auxiliary term at a time on the same parents and improves each term's target on all five. At 1,200 pretraining updates, composition alone increases IID task error on all five ($+13.7\%$ geometric-mean change); anchoring increases it on only two ($-0.6\%$), with a descriptive interval $[-11.7\%,+11.8\%]$ permitting material effects in either direction (\cref{fig:synthetic-diagnostics}B). This interval does not establish that anchoring has no effect. The same distinction in the number of parents harmed holds at 4,800 updates and under distribution shift. \Cref{app:factor-decomp} gives the exact comparisons and their scope.

\paragraph{Robustness and limits of the attempted remedy.}
The candidate exceeds its own fine path's loss by more than the allowed $2\%$ in all six families (\cref{fig:family-quality}). Increasing fine pretraining up to $16\times$ improves some fine references but never meets the prespecified criterion against the stronger DIRECT controls; improvements occur only relative to a cheaper adaptation-only control. Chaining the uncalibrated auxiliary maps for single-update spans also raises task loss relative to the macro replacing both fine updates. Varying both auxiliary weights together does not isolate either term's effect. The benefit of response matching, and the relation between fidelity and task error, vary across families and budgets. Protocols and results are in \cref{app:multifamily,app:teacher-regime,app:factor-decomp}; detailed diagnostics are in the numerical supplement.
\subsection{Task quality and complete deployment cost}\label{fmt:utility}
Always using the macro is $2.22\times$ faster than its own fine path on the measured CPU, but increases IID MSE by $42.19\%$. Although its MSE is lower than that of continued fine-only training, it fails the joint $2\%$ quality rule. The speedup therefore comes with a loss of quality beyond the allowed tolerance.

After the paths are frozen, an input-only selector predicts their losses and chooses a path using a threshold fixed on development data. The primary two-layer MLP and a linear control are trained on path losses computed offline on the training split, with the cost of generating these targets accounted for. Measured latency includes selector prediction, partitioning examples by path, and dispatch. In addition to meeting both fine-reference quality criteria, selection on IID data must save at least $10\%$ of affine MACs and take at most $90\%$ of the candidate's own fine-path latency. \Cref{app:synthetic-selection} gives training and aggregation details.

A development-set oracle that uses labels permits $43.01\%$ affine-work saving under its specified constraint. This offline bound uses targets unavailable to input-only selectors and does not establish attainable test performance. \Cref{tab:unified-dynamic} reports the deployment comparison.

\begin{table}[!htbp]
\centering\small
\caption{\textbf{Neither input-only selector meets the joint quality--cost requirements.} IID test results over five parents, relative to the candidate's own fine path. Both MSE ratios exceed $1.02$ and therefore fail the quality rule, which requires meeting the tolerance against both fine references. Column headers give the required thresholds: smaller MSE and latency ratios and larger MAC savings are better. Latency includes selector evaluation, partitioning, and dispatch; affine MAC savings include selector work but exclude non-affine overhead. Ratios use means across parents, with median latency measured per parent.}
\label{tab:unified-dynamic}
\begin{tabular}{@{}lrrr@{}}
\toprule
Selector & MSE/fine $\le1.02$ & MAC saving $\ge10\%$ & Latency/fine $\le0.90$ \\
\midrule
MLP (primary) & 1.0870 & $9.71\%$ & 1.2378 \\
Linear (control) & 1.1511 & $19.73\%$ & 0.8968 \\
\bottomrule
\end{tabular}
\end{table}

\begin{samepage}
Relative to the candidate's own fine path, the MLP reduces affine work but runs more slowly; the linear selector runs faster but has higher task error. Both exceed the quality tolerance under distribution shift. No tested primary selector policy meets aggregate quality on either split. Latency under shift and separate selection and dispatch overheads were not measured. Curves with thresholds fixed on development data describe tested operating points; they do not establish an untested point that meets the quality requirement.
\par\end{samepage}
\FloatBarrier
\section{Discussion, Limitations, and Conclusion}\label{fmt:discussion}
\paragraph{What the evidence establishes.}
The framework evaluates retained information through a specified continuation, task loss, and complete cost. The finite construction shows lower assigned cost under explicit restrictions on information and controllers. The pretrained-model studies show that local fidelity and task quality can rank substitutes differently, with rankings changing across distributions. The recursive controls show why a favorable diagnostic or a faster substitute alone does not establish useful replacement. These findings support evaluating quality and cost together, building on established distinctions between fidelity and usefulness.

The candidate and DIRECT controls use the same deployed graph, so differences cannot be attributed to greater expressivity. To outperform these controls, a training procedure would need to improve learning at a fixed budget, robustness, or another specified objective. Here, adapting more components helps relative to restricted adaptation, but does not beat joint boundary-state training or DIRECT. The single-term tests also show why the combined objective's harm cannot be attributed to fidelity matching itself: composition and occupancy anchoring have different effects. The exact gap identity describes frozen-model behavior without identifying its training causes.

\paragraph{Demonstrated scope.}
The studies cover related synthetic families with five parents each, a width-16 eight-expert prototype, and CPU timing at batch size one, plus two dense pretrained checkpoints and one affine substitute class. Replacement boundaries share the same model, and fitting-order seeds are not independent model replications. The larger-model study uses 21 fitting articles in source order and one update budget; its article bootstrap conditions on the fitted substitutes. Neither language-model study includes an equally adapted full-model control or a substitute that matches or beats the unadapted full model on every held-out set. Matched updates and forward affine operations do not match backward-pass or optimizer cost. Interrupted language-model timing is descriptive. Incomplete records of historical model and input versions, and missing arrays, limit reproducibility (\cref{app:realmodel,app:qwen3,app:reproducibility-details}).

The formal interface requires explicit retained history and consistent action meanings. Larger sparse expert banks, arbitrary-horizon control, robot policies, and generative flow matching remain untested; side effects, asynchronous tools, closed-loop behavior, and routing side channels \citep{ding2025moecho} require additional analysis. These limits do not imply that useful compression is impossible. The remaining challenge is to learn fine computation, replacements, and efficient path selection that jointly meet the evaluation protocol's quality and cost requirements.
\paragraph{Conclusion.}\label{fmt:conclusion}
We formulate a framework for evaluating computational replacements through their continuation, supported by conditional analysis, an exactly enumerated construction with restricted controllers, and controlled comparisons. Local agreement, task quality, and execution savings require separate measurements. The observed objective rankings depend on the evaluation distribution, and the recursive candidate does not achieve quality-matched acceleration. Useful compression must therefore be demonstrated through the specified continuation, quality references, and complete cost; local fidelity alone is insufficient.
\label{fmt:main-end}
\FloatBarrier

\subsection*{AI use statement}
\label{fmt:ai-statement}
Generative AI tools (chat assistants and coding agents) assisted in developing the framework and proofs, designing and running experiments, implementing code, analyzing results, and preparing the manuscript and figures. Verification of proofs, numerical results, and citations also involved AI assistance. The authors directed and reviewed the work and take full responsibility for its content.

\subsection*{Reproducibility statement}
\label{fmt:reproducibility}
The Supplementary Information (SI) provides derivations, protocols, and task-loss comparisons for the reported experimental conditions. The source archive includes a numerical supplement with the metrics underlying the condensed GPT-2 and Qwen3 displays. Private author-review packages contain code and verification tools; for Qwen3, they also contain version-identified inputs, per-example metric sums, and substitute checkpoints. Some historical arrays and checkpoints are unavailable, and GPT-2 model and input revisions and one fitting-corpus source remain unresolved. Public release and licensing of the complete experimental evidence remain pending (\cref{app:reproducibility-details,app:realmodel,app:qwen3}).

\FloatBarrier
\label{fmt:references-start}
\begingroup
\setlength{\bibsep}{6pt plus 1pt minus 1pt}
\bibliographystyle{plainnat}
\bibliography{references}
\endgroup

\clearpage
\appendix
\setcounter{equation}{0}
\setcounter{table}{0}
\setcounter{figure}{0}
\setcounter{proposition}{0}
\setcounter{theorem}{0}
\setcounter{lemma}{0}
\setcounter{definition}{0}
\setcounter{remark}{0}
\renewcommand{\theequation}{S\arabic{equation}}
\renewcommand{\thetable}{S\arabic{table}}
\renewcommand{\thefigure}{S\arabic{figure}}
\renewcommand{\theproposition}{S\arabic{proposition}}
\renewcommand{\thetheorem}{S\arabic{theorem}}
\renewcommand{\thelemma}{S\arabic{lemma}}
\renewcommand{\thedefinition}{S\arabic{definition}}
\renewcommand{\theremark}{S\arabic{remark}}
\renewcommand{\theHequation}{S.\arabic{equation}}
\renewcommand{\theHtable}{S.\arabic{table}}
\renewcommand{\theHfigure}{S.\arabic{figure}}
\providecommand{\theHproposition}{\arabic{proposition}}
\renewcommand{\theHproposition}{S.\arabic{proposition}}
\providecommand{\theHlemma}{\arabic{lemma}}
\renewcommand{\theHlemma}{S.\arabic{lemma}}
\section*{Supplementary Information}
\label{fmt:si-start}
The SI explains the interface through short derivations with explicit assumptions, followed by methods and complete comparisons for the finite construction, synthetic experiments, and two pretrained-language-model studies. Equations, tables, and figures carry an S prefix. The evaluated populations, budgets, and aggregation conventions are the same as in the main text.

\paragraph{Reader's guide.}
\Cref{si:formal} develops the intuition, accounting rules, and approximation limits; \cref{app:scale-k-r-sealed} specifies the exact finite construction; \cref{si:synthetic} describes the eight-expert studies; and \cref{app:realmodel} presents the language-model protocols and complete task-loss comparisons, with the larger-model replication in \cref{app:qwen3}. Additional exploratory records and figures are included in the numerical supplement. Essential methods, derivations, and unfavorable outcomes remain in the SI; \cref{app:reproducibility-details} describes the availability and provenance of the evidence.
\section{Understanding the Interface and Its Limits}\label{si:formal}
\subsection{What Does a Coarse Step Retain?}
\label{fmt:appendix-start}

The definitions below distinguish the state passed to later computation, statistics of skipped steps, and operations actually executed. The equations specify model inputs, outputs, and computation costs without treating neural computation as a physical system.

\subsubsection{The fine path and its effective update}
The retained routing statistics of a computational path form its \emph{routing summary}. Let $c\in\mathcal C$ denote external conditioning, $d,E,K$ be positive integers, and normalized computation time be $s\in[0,1]$. Write $\simplex=\{p\in\R^E:p_e\ge0,\ \bm1^\top p=1\}$ for the probability simplex of $E$ expert weights. With $E$ experts, the fine path (the \emph{teacher}) uses $K$ updates of width $h=1/K$:
\begin{align}
 z_{i+1} &= F_i(z_i,c,b_i), \nonumber\\
 \pi_i &= R_i(z_i,c,b_i)\in\simplex,
 \qquad i=0,\ldots,K-1. \label{eq:fine}
\end{align}
Here $z_i\in\mathcal Z\subseteq\R^d$ is the visible state, $\pi_i$ contains soft expert responsibilities (weights), and $b_i$ is any history required for \emph{closure}: the augmented state, context, and step index determine the next update without additional incoming history. A memoryless router omits $b_i$. For a hard top-$k$ selection, with integer $1\le k\le E$, we record activations by a normalized multi-hot vector $\alpha_i$: each selected coordinate is $1/k$ and each other coordinate is zero. Capacity drops without renormalization instead give a sub-probability vector with dropped mass $d_i=1-\bm 1^\top\alpha_i\ge0$.

When $z$ is a sufficient state, the teacher transition over an interval is
\begin{equation}
  \teacher_{s,\Delta}(z;c)=F_{j+m-1}\circ\cdots\circ F_j(z;c),
  \quad s=jh,\quad \Delta=mh. \label{eq:teacher-map}
\end{equation}
Here $j\ge0$ and $m\ge1$ are integers with $j+m\le K$. Because $F_i$ may depend on $i$, $\teacher$ depends on both the starting depth and the interval length: it is a two-parameter process. It satisfies
\begin{equation}
  \teacher_{s,\Delta_1+\Delta_2}
  =\teacher_{s+\Delta_1,\Delta_2}\circ\teacher_{s,\Delta_1}, \label{eq:process-law}
\end{equation}
for positive, grid-aligned $\Delta_1,\Delta_2$ with $s+\Delta_1+\Delta_2\le1$. A one-parameter semigroup is the special case of autonomous dynamics. If $z$ is not sufficient, \cref{eq:teacher-map,eq:process-law} apply instead to the augmented state $x=(z,b)$ and its closed updates, including the update of $b$; \cref{prop:closure} quantifies why this distinction matters. The continuous idealization below uses valid positive intervals; it does not assert that the discrete teacher is defined between grid points.

\subsubsection{Averages retain amounts, not order}\label{prop:order}
For a nonempty discrete interval $I$ containing $n\ge1$ route vectors $p_0,\ldots,p_{n-1}$, where $p_i$ can be either $\pi_i$ or $\alpha_i$, define total routing mass and step-averaged occupancy:
\begin{equation}
 A_I=\sum_{i=0}^{n-1}p_i,
 \qquad Q_I=A_I/n. \label{eq:occupancy-discrete}
\end{equation}
For soft responsibilities, $Q_I\in\simplex$. With recorded capacity drops it is instead a sub-probability vector, with total mass $1-D_I/n$ for $D_I$ defined below. Here $A_I$ sums activation mass over steps; for width $h$ and duration $\Delta=nh$, the corresponding mass integrated over duration is $hA_I=\Delta Q_I$. For a measurable continuous probability-valued responsibility path on $[s,s+\Delta]$, with $\Delta>0$, the duration-normalized occupancy is
\begin{equation}
 Q^\star_{s,\Delta}(z;c)
 =\frac{1}{\Delta}\int_s^{s+\Delta}\pi_r(z_r,c)\,dr\in\simplex. \label{eq:occupancy}
\end{equation}
In the continuous idealization, assume a closed evolution and use $\teacher_{s,\Delta}(z;c)$ for its state at $s+\Delta$ when started from $z$ at $s$. In the discrete setting, use the grid-aligned map in \cref{eq:teacher-map} and step-averaged occupancy. In either setting, the accumulated mass $\Delta Q^\star$ satisfies an additive \emph{cocycle} identity; equivalently, occupancy composes by duration-weighted averaging:
\begin{align}
 &(\Delta_1+\Delta_2)Q^\star_{s,\Delta_1+\Delta_2}(z;c) \nonumber\\
 &\quad=\Delta_1Q^\star_{s,\Delta_1}(z;c)
 +\Delta_2Q^\star_{s+\Delta_1,\Delta_2}
 \bigl(\teacher_{s,\Delta_1}(z;c);c\bigr). \label{eq:cocycle}
\end{align}
The successor state in the second term matters because routing depends on the state. A cocycle describes how a quantity accumulates along an evolving state: intervals contribute in proportion to their lengths, and the second interval begins where the first ends.

An average erases chronology. For $A(z)=2z$ and $B(z)=z+1$, executing each once gives occupancy $(1/2,1/2)$, but $B(A(z))=2z+1$ whereas $A(B(z))=2z+2$. The difference comes from operator order, not from the amount assigned to either expert.

The occupancy summary is $S_I^{(0)}=(n,A_I)$. An optional adjacent-pair summary $S_I^{(1)}$ additionally records consecutive routing pairs and the first and last routes. The superscripts distinguish these two summaries:
\begin{equation}
 \begin{aligned}
 S_I^{(1)}&=(n,A_I,B_I,\ell_I,r_I,D_I),\\
 B_I&=\sum_{i=0}^{n-2}p_ip_{i+1}^{\top},\\
 (\ell_I,r_I)&=(p_0,p_{n-1}),
 \end{aligned}
 \label{eq:ordered-signature}
\end{equation}
where $D_I=\sum_i(1-\bm1^\top p_i)$ records dropped mass. The empty sum gives $B_I=0$ for a singleton interval. If nonempty adjacent intervals $I$ and $J$ are concatenated,
\begin{align}
 n_{IJ}&=n_I+n_J, & A_{IJ}&=A_I+A_J,\nonumber\\
 D_{IJ}&=D_I+D_J, &
 (\ell_{IJ},r_{IJ})&=(\ell_I,r_J),\nonumber\\
 B_{IJ}&=B_I+B_J+r_I\ell_J^\top. \label{eq:ordered-compose}
\end{align}
The outer product accounts for the adjacent pair across the boundary. Define $C_I=B_I/(n-1)$ for $n>1$, and set $C_I=0$ for $n=1$ by convention; a singleton has no observed adjacent pair. Retaining $n_I$ lets us compose the normalized transition frequencies as
\begin{equation}
 C_{IJ}=\frac{(n_I-1)C_I+(n_J-1)C_J+r_I\ell_J^\top}{n_I+n_J-1}. \label{eq:transition-cocycle}
\end{equation}
Every adjacent pair lies inside $I$, inside $J$, or across their boundary; this gives \cref{eq:ordered-compose}. Three joined intervals add the same two boundary terms in either grouping, so the rule is associative. Empty intervals are excluded because their first and last routes are undefined. The prototype uses $S^{(0)}$; $S^{(1)}$ is an algebraic extension with code for checking its identities, not a trained prediction head. It distinguishes the two-operation example but need not identify an arbitrary long operator sequence.

\subsubsection{A path summary is not automatically a memory state}
For sufficient $z_s$, a coarse interval model produces an endpoint and a path observable,
\begin{equation}
 (z_s,c,s,\Delta)\longmapsto
 \bigl(z_{s+\Delta},S_{[s,s+\Delta]}\bigr). \label{eq:observable-model}
\end{equation}
The routing summary describes an interval. If routing history at the interval boundary, $b_s$, carries predictive information beyond $z_s$, closure instead requires
\begin{equation}
 (z_s,b_s,c,s,\Delta)\longmapsto
 \bigl(z_{s+\Delta},b_{s+\Delta},S_{[s,s+\Delta]}\bigr). \label{eq:product-state-model}
\end{equation}
Only \cref{eq:product-state-model} propagates $b_{s+\Delta}$ as part of the dynamic state. Conditioning on occupancy, as the prototype does, does not establish closure for arbitrary histories or replacement schedules.

\subsubsection{Predicted information and executed work}
For admissible $(s,\Delta)$, coarse models with shared parameters predict
\begin{equation}
 \begin{aligned}
 \widehat z_{s+\Delta}&=\student(z_s,c,s,\Delta),\\
 \widehat S_{s,\Delta}&=\widehat S_\theta(z_s,c,s,\Delta).
 \end{aligned}
 \label{eq:student}
\end{equation}
A separate execution route $\rho_\theta$ chooses which coarse experts run. A positive integer budget $N$ specifies an externally chosen partition into $N$ positive intervals summing to one; grid alignment implies $N\le K$. Any required routing memory is added to the input and output as in \cref{eq:product-state-model}. The prototype selects between two fixed paths at one boundary, rather than learning $N$; total execution cost is evaluated in \cref{fmt:utility}.

\subsubsection{Which clock and capacity are being changed?}\label{sec:three-clocks}
Physical control time, generative-flow time within one prediction, and internal recursive depth are distinct; compressing the specified recursive coordinate does not compress the others. Stored parameter capacity, the capacity activated on each update, and effective depth from weight reuse are likewise distinct.

\begin{table}[htbp]
\centering
\small
\caption{Notation for the routing formulation in \cref{fmt:appendix-start,app:signature-formulation}. Notation for continuation actions and losses is defined separately in \cref{sec:continuation-validity}.}
\label{tab:notation}
\begin{tabularx}{0.98\linewidth}{@{}lX@{}}
\toprule
Symbol & Meaning \\
\midrule
$c$; $s,\Delta$ & external context; start and length of the computation interval \\
$z_s,b_s$ & visible state; optional history state needed for closure \\
$E;\pi_s;\alpha_s;d_s$ & expert count; soft responsibilities; hard activations; dropped mass \\
$\teacher_{s,\Delta};\student$ & fine teacher transition; learned coarse state map \\
$A_I,Q_I$ & unnormalized route mass; normalized occupancy \\
$B_I,C_I,\ell_I,r_I$ & mass and frequency of adjacent routing pairs; first and last route \\
$\widehat S_\theta;\rho_\theta$ & predicted teacher-path summary; actual student execution route \\
$\delta_z,\delta_q$ & state- and occupancy-composition defects \\
$N,k$ & deployed model evaluations; active experts per coarse evaluation \\
\bottomrule
\end{tabularx}
\end{table}
\subsection{What Do Fidelity and Consistency Losses Measure?}\label{app:signature-formulation}

These losses measure fidelity to the teacher and consistency across interval lengths for \cref{eq:student}, without specifying an additional architecture. Actual execution routing remains distinct from predicted teacher occupancy.

For occupancy summaries, $\widehat Q_\theta$ denotes the predicted normalized occupancy. In the bounds, $Q_{\theta;s,\Delta}(z)$ abbreviates $\widehat Q_\theta(z,c,s,\Delta)$ with conditioning $c$ suppressed; $\Phi_{\theta;s,\Delta}$ similarly abbreviates the state map.

\subsubsection{Agreement with the fine path}
For equal-dimensional vectors, write $\operatorname{mse}(v,w)=\norm{v-w}_2^2/\dim(v)$. For a sampled teacher interval $I=(z_s,c,s,\Delta)$, define
\begin{align}
 \mathcal L_{\mathrm{jump}}
 &=\E_I\!\left[\operatorname{mse}\!\left(\student(I),\teacher_{s,\Delta}(z_s;c)\right)\right],\label{eq:jump-loss}\\
 \mathcal L_{Q\text{-}\mathrm{anchor}}
 &=\E_I\!\left[D_A\!\left(\widehat Q_\theta(I),Q^\star_{s,\Delta}(z_s;c)\right)\right]. \label{eq:route-loss}
\end{align}
Here $\E_I$ averages over the chosen distribution of training examples and intervals. The prototype samples training examples uniformly with replacement and fixes the interval to the complete two-update prefix. The discrepancy $D_A$ compares predicted occupancy with the target, in that order. Comparing coordinates requires shared expert identities or the fixed calibration alignment in \cref{eq:fixed-alignment}. The prototype uses coordinate MSE and stops gradients through the current fine-path occupancy target.

\subsubsection{Agreement between a long step and two short steps}
For a valid partition $\Delta=\Delta_1+\Delta_2$, define
\begin{align}
 \widehat z_1 &=\student(z,c,s,\Delta_1),\\
 \widehat z_{12}^{\mathrm{comp}}
 &=\student(\widehat z_1,c,s+\Delta_1,\Delta_2),\\
 \widehat Q_{12}^{\mathrm{comp}}
 &=\frac{\Delta_1}{\Delta}\widehat Q_\theta(z,c,s,\Delta_1)\nonumber\\
 &\quad+\frac{\Delta_2}{\Delta}
 \widehat Q_\theta(\widehat z_1,c,s+\Delta_1,\Delta_2).
\end{align}
The corresponding penalties also average over the choice of valid partition; the prototype fixes that partition to two equal intervals:
\begin{align}
 \mathcal L_{\mathrm{state\text{-}comp}}
 &=\E\operatorname{mse}\!\left(\student(z,c,s,\Delta),\widehat z_{12}^{\mathrm{comp}}\right),\label{eq:state-comp-loss}\\
 \mathcal L_{Q\text{-}\mathrm{cocycle}}
 &=\E D_C\!\left(\widehat Q_\theta(z,c,s,\Delta),\widehat Q_{12}^{\mathrm{comp}}\right),\label{eq:route-comp-loss}
\end{align}
with $D_C=\operatorname{mse}$ for the prototype. A constant occupancy satisfies the weighted composition rule, so consistency alone cannot identify the teacher target. Mean squared errors on sampled intervals do not establish the pointwise norm and total-variation assumptions of \cref{prop:state-bound,prop:route-bound}.

The prototype's composition and response penalties propagate gradients through both operands. Its span arguments count fine updates: with $K=2$, a start index $j$ and span $n$ correspond to $s=j/2$ and $\Delta=n/2$. Only the two-update span is directly trained to match the fine path; the one-update maps are auxiliary factors, not calibrated teacher steps. See \cref{app:synthetic-model,sec:unified-v3} for the full objective, parameter sharing, and separately fitted selector.
\subsection{What Information and Errors Survive Coarse-Graining?}
\label{app:signature-analysis}\label{app:proofs}
These elementary calculations clarify what the experiments can establish; we do not claim them as new general theorems. State norms are fixed, Lipschitz constants use that norm, and routing discrepancy is total variation $D(q,p)=\tfrac12\norm{q-p}_1$. All bounds are finite and nonnegative. The maps must be defined and the bounds must hold on the relevant true \emph{and predicted} states; sampled mean squared errors do not establish these uniform assumptions.

\subsubsection{When does the coarse state need memory?}\label{prop:closure}
A useful physical analogy is eliminating microscopic degrees of freedom: a retained variable forms a closed state only if it determines the next update. Suppose two allowed histories have the same visible input $(z,c,s,\Delta)$ but lead to endpoints $y,y'$ separated by $d_{\rm sep}=\norm{y-y'}>0$. Any deterministic prediction based only on that input, $\widehat y=\chi(z,c,s,\Delta)$, is the same for both. Since
\[
d_{\rm sep}\le\norm{y-\widehat y}+\norm{\widehat y-y'},
\]
at least one error is $d_{\rm sep}/2$ or larger. Fitting the same input more accurately cannot resolve this ambiguity: the relevant memory must be included in the propagated state. A distributional predictor can represent the uncertainty without identifying which hidden history occurred. The tested memoryless prefix has no ambiguity from omitted history, so this calculation does not explain its observed failure.

For a continuous idealization in the computation coordinate $r$, suppose an absolutely continuous state obeys $\dot z_r=\sum_e\pi_e(r)f_e(z_r,c,r)$ almost everywhere. With measurable probability-valued $\pi$ and integrable weighted fields over $\Delta>0$,
\begin{equation}
\begin{aligned}
z_{s+\Delta}-z_s&=\Delta\sum_e Q_e^\star\,\bar f_e,\\
\bar f_e&=\frac{\int_s^{s+\Delta}\pi_e(r)f_e(z_r,c,r)\,dr}
{\int_s^{s+\Delta}\pi_e(r)\,dr}.
\end{aligned}
\label{eq:phase-decomp}
\end{equation}
Set $\bar f_e=0$ when its denominator vanishes. This identity regroups the integrated update exactly. Occupancy records how much weight each expert receives; the conditional average $\bar f_e$ also depends on the states visited. Occupancy alone therefore does not determine the effective update. This idealization asserts neither physical conservation nor that the discrete teacher is defined between grid points.

\subsubsection{How do local errors grow across coarse steps?}\label{prop:state-bound}\label{prop:sequential}
A composition loss tests whether one long learned step agrees with two short learned steps; it does not test their agreement with the fine computation. Let $\widehat z_1=\Phi_{\theta;s,\Delta_1}(z)$ and $z_1^\star=\teacher_{s,\Delta_1}(z)$ for valid positive intervals. Suppose the discrepancy between the long and composed learned steps is at most $\delta_z$, errors relative to the teacher on both subintervals are uniformly bounded by $\varepsilon_1,\varepsilon_2$, and the second teacher map is $L_2$-Lipschitz on the relevant states. Add and subtract the learned composition and the teacher evaluated at $\widehat z_1$:
\begin{equation}
\begin{aligned}
&\norm{\Phi_{\theta;s,\Delta_1+\Delta_2}(z)-\teacher_{s,\Delta_1+\Delta_2}(z)}\\
&\ \le
\underbrace{\norm{\Phi_{\theta;s,\Delta_1+\Delta_2}(z)
-\Phi_{\theta;s+\Delta_1,\Delta_2}(\widehat z_1)}}_{\text{composition defect}\ \le\delta_z}\\
&\qquad+\underbrace{\norm{\Phi_{\theta;s+\Delta_1,\Delta_2}(\widehat z_1)
-\teacher_{s+\Delta_1,\Delta_2}(\widehat z_1)}}_{\text{local approximation}\ \le\varepsilon_2}\\
&\qquad+\underbrace{\norm{\teacher_{s+\Delta_1,\Delta_2}(\widehat z_1)
-\teacher_{s+\Delta_1,\Delta_2}(z_1^\star)}}_{\text{propagated state error}\ \le L_2\varepsilon_1}\\
&\ \le\delta_z+\varepsilon_2+L_2\varepsilon_1 .
\end{aligned}
\label{eq:state-bound}
\end{equation}
The process law in \cref{eq:process-law} identifies the teacher endpoint. The local approximation bound must hold at the \emph{predicted} successor, not just on teacher trajectories.

For $N$ sequential intervals, let $e_j=\norm{\widehat z_j-z_j^\star}$ denote the state discrepancy at the $j$th coarse boundary. With a common initial state, uniform local approximation error relative to the teacher $\varepsilon$, and teacher Lipschitz constant $L$ on the reached states, the same argument gives $e_{j+1}\le\varepsilon+Le_j$, $e_0=0$. Thus
\[
e_N\le\varepsilon\sum_{j=0}^{N-1}L^j
=\begin{cases}\varepsilon N,&L=1,\\
\varepsilon(L^N-1)/(L-1),&L\ne1.
\end{cases}
\]
This is the familiar amplification of a local error by subsequent dynamics: contraction ($L<1$) limits accumulation, whereas the bound can grow rapidly for $L>1$. Neither a small composition loss nor a short sequence of deployed steps alone establishes the required accuracy or stability.

\subsubsection{How does state error enter a route summary?}\label{prop:route-bound}\label{prop:mixture}
A routing summary is evaluated along a trajectory. Changing that trajectory can change the summary even when its predictor is locally accurate. Let $\Delta=\Delta_1+\Delta_2$ for valid positive subintervals $\Delta_1,\Delta_2$. Suppose the learned occupancy violates the weighted composition rule by at most $\delta_q$, the two occupancy prediction errors at true states are at most $\eta_1,\eta_2$, and the second occupancy predictor is $K_q$-Lipschitz under $D$ between the true and predicted successor. If their state discrepancy is at most $\varepsilon_1$, weighting by interval length and applying the triangle inequality give
\begin{equation}
D(Q_{\theta;s,\Delta},Q^\star_{s,\Delta})
\le\delta_q+
\frac{\Delta_1\eta_1+\Delta_2(\eta_2+K_q\varepsilon_1)}
{\Delta_1+\Delta_2}.
\label{eq:route-bound}
\end{equation}
The additional $K_q\varepsilon_1$ accounts for evaluating the second predictor at the predicted successor. It follows by inserting its prediction at the true successor into the triangle inequality. This derivation uses probability-valued summaries and the exact weighted composition law, not an unweighted average of unequal intervals.

Conversely, at a \emph{common} state with unchanged expert updates satisfying $\norm{f_e}\le M$,
\begin{equation}
\norm{\sum_e(q_e-p_e)f_e}
\le\sum_e|q_e-p_e|\norm{f_e}
\le2M D(q,p).
\label{eq:mixture-bound}
\end{equation}
If the same bound $M$ holds throughout a common path of states and expert fields, integrating gives at most $2M\Delta$ times the interval's mean total-variation error. This isolates the effect of changed mixture weights; different state trajectories or expert functions require additional terms. It does not make occupancy matching a guarantee of task quality.

\subsubsection{Why can different expert allocations look identical?}\label{lem:identifiability}
At a fixed input, collect expert outputs as columns of $V\in\R^{d\times E}$. If a nonzero vector $a$ satisfies $Va=0$ and $\bm1^\top a=0$, then changing the weights from $q$ to $q+ta$ leaves both their sum and the mixture output unchanged. For $q\in\simplex$ with every $q_e>0$, sufficiently small positive and negative $t$ preserve nonnegativity, giving infinitely many nearby allocations with the same output. Such a direction is guaranteed when $E>d+1$, because the augmented matrix $[V^\top,\bm1]^\top$ has at most $d+1$ independent rows. The general condition is the stated nullspace condition, not expert count alone.

Thus matching endpoints need not recover the teacher's specified responsibilities. Conversely, a useful replacement need not reproduce one unique expert decomposition. Even an exact occupancy loses ordering information, as the two-operation example in \cref{prop:order} shows.

\paragraph{Keeping expert coordinates comparable.}
A shared expert bank fixes the meaning of each coordinate; separately trained banks may change expert functions or permute their identities. If the evaluation protocol allows a permutation, choose it once on a disjoint nonempty calibration set:
\begin{equation}
\begin{aligned}
P^\star&\in\operatorname*{arg\,min}_{P\in\mathcal P_E}
\frac{1}{|\mathcal D_{\rm cal}|}\sum_{i\in\mathcal D_{\rm cal}}D(q_i,Pp_i),\\
D_{\rm aligned}(q,p)&=D(q,P^\star p),
\end{aligned}
\label{eq:fixed-alignment}
\end{equation}
where $\mathcal P_E$ is the set of permutation matrices. Fix the objective and deterministic tie rule before testing, then use the same $P^\star$ across examples and schedules. The prototype uses a shared bank and fits no alignment. Relabeling coordinates does not establish functional equivalence of the continuations.

\subsubsection{When does a small perturbation change executed experts?}\label{prop:topk}\label{prop:support-floor}
Hard selection is discontinuous at a ranking tie. For teacher and replacement logits $g,\widehat g\in\R^E$ and $1\le k<E$, let $\gamma>0$ be the gap between the $k$th and $(k+1)$st teacher logits. If $\norm{\widehat g-g}_\infty<\gamma/2$, every selected $i$ and unselected $j$ satisfy
\[
\widehat g_i-\widehat g_j
\ge g_i-g_j-2\norm{\widehat g-g}_\infty>0.
\]
Their top-$k$ sets therefore agree before capacity limits are applied. For $k=E$ the whole bank is selected without a margin condition. Equality at the threshold permits ties; capacity limits can still change which experts execute. Interval-averaged occupancy does not establish this pointwise logit condition.

A separate counting argument limits how many distinct experts can execute. In $N$ macrosteps selecting at most $k$ experts each, at most $m=\min(E,Nk)$ distinct experts can run. For a probability-valued teacher occupancy $q$ in the same expert coordinates, sorted as $q_{(1)}\ge\cdots\ge q_{(E)}$,
\begin{equation}
\min_{p\in\simplex:\,|\operatorname{supp}(p)|\le m}D(p,q)
=1-\sum_{e=1}^{m}q_{(e)}.
\label{eq:support-floor}
\end{equation}
To see this, a retained support $S$ omits mass $r=\sum_{e\notin S}q_e$. The absolute differences outside the support sum to $r$; normalization forces their sum inside the support to be at least $r$, so total variation is at least $r$. Retaining the $m$ largest entries minimizes it. Keeping their original weights and adding the omitted mass to one retained coordinate attains equality in this relaxed class of probability vectors. An actual routing schedule need not attain it. Capacity drops require accounting for sub-probability mass. The limit applies to \emph{executed} occupancy: a dense predicted summary can assign weight to experts that never run, and this bound says nothing by itself about task loss.
\subsection{From Local Fidelity to Downstream Task Loss}
\label{sec:continuation-validity}
Subsequent computation can both amplify a replacement error and respond to it by choosing a different action. Separating these effects explains why a locally faithful substitute need not give the best task performance.

\subsubsection{What is held fixed in this comparison?}
For one input--target pair $(x,y)$ and frozen parameters, let $\vartheta$ collect the shared encoder, experts, router and readout, and $\varphi$ the macro parameters. These symbols are distinct from $\theta$ used by the interval model above. With $u_0=\operatorname{Enc}_{\vartheta}(x)$,
\begin{equation}
u^F=\mathcal F_{\vartheta}(u_0),\qquad
u^M=\mathcal C_{\varphi}(u_0).
\label{eq:continuation-states}
\end{equation}
The fine prefix and macro reach the same logical stage. For a common finite nonempty action set $\mathcal A$, $o_j(a)$ is the shared readout's output after action $a$ from $u^j$, $j\in\{F,M\}$. The continuation maps and router are shared deterministic functions. The router selects $a_j$ from its input, without seeing $y$ or all candidate outputs. Fix its tie rule. Actions must retain the same meaning across paths, or use a specified alignment with no rematching for individual test examples. Such equivalence is relative to a specified policy or query class \citep{grimm2020value,chen2026cqm}.

Define squared task loss, the minimum loss over available actions, and router regret (excess loss over that minimum):
\begin{equation}
\ell_j(a)=\tfrac12\norm{o_j(a)-y}_2^2,\qquad
m_j=\min_{a\in\mathcal A}\ell_j(a),\qquad
\mathfrak r_j=\ell_j(a_j)-m_j\ge0.
\label{eq:continuation-risk}
\end{equation}
The minimum is an offline diagnostic computed using the target, which the deployment policy cannot access. For output dimension $d$, this loss is $d/2$ times coordinate MSE; it equals MSE in the two-output prototype.

\subsubsection{What changed: the substitute or the next decision?}\label{prop:continuation-standard}
First keep the fine action fixed while changing the prefix; then change the action on the macro path. Adding and subtracting the intermediate loss gives
\begin{align}
\ell_M(a_M)-\ell_F(a_F)
&=\underbrace{\ell_M(a_F)-\ell_F(a_F)}_{\text{fixed-action replacement}}\nonumber\\
&\quad+\underbrace{\ell_M(a_M)-\ell_M(a_F)}_{\text{reallocation}}
\label{eq:replacement-reallocation}\\
&=m_M-m_F+\mathfrak r_M-\mathfrak r_F .
\label{eq:continuation-regret}
\end{align}
A negative reallocation term lowers loss relative to keeping the fine action fixed; a positive one raises it. This identity compares paths at fixed parameters and does not identify how training caused the result. The reference choice matters: reversing the reference gives the equally exact decomposition
$[\ell_M(a_M)-\ell_F(a_M)]+[\ell_F(a_M)-\ell_F(a_F)]$.
The experiments use the fine-action reference in \cref{eq:replacement-reallocation} (\cref{app:decomp-shift}).

\paragraph{What does uniform fidelity actually control?}
Suppose losses are finite and $|\ell_M(a)-\ell_F(a)|\le\varepsilon_\ell$ for \emph{every} action. Testing a fine minimizer on the macro path gives $m_M\le m_F+\varepsilon_\ell$; exchanging the paths gives the reverse bound. Consequently
\begin{equation}
\left|\ell_M(a_M)-\ell_F(a_F)-(\mathfrak r_M-\mathfrak r_F)\right|
\le\varepsilon_\ell .
\label{eq:continuation-bound}
\end{equation}
Fidelity controls the change in the best available loss, but the change in router regret remains. For Euclidean output errors $e_a=o_M(a)-o_F(a)$, the identity
\[
\ell_M(a)-\ell_F(a)
=\langle o_F(a)-y,e_a\rangle+\tfrac12\norm{e_a}_2^2
\]
makes the required scale explicit: if $\norm{e_a}_2\le\varepsilon_o$ and $\norm{o_F(a)-y}_2\le B_{\rm res}$ uniformly, Cauchy--Schwarz gives $\varepsilon_\ell=B_{\rm res}\varepsilon_o+\varepsilon_o^2/2$, with finite nonnegative bounds.

The fixed-action term is also bounded by $\varepsilon_\ell$. Thus choosing the same action on both paths implies a total gap of at most $\varepsilon_\ell$ in magnitude. Exact equality of \emph{every input used by the continuation}, including history, context and occupancy, under the shared maps and tie rule makes outputs and actions identical and the gap zero. Equality of selected statistics or even all candidate responses alone does not ensure identical router inputs and choices.

Subtracting the fixed-action term from \cref{eq:continuation-bound} further bounds the difference between reallocation and regret change by $2\varepsilon_\ell$. It does not bound either quantity alone. A response loss averaged over a minibatch establishes neither these assumptions for every action nor a regret bound.

\paragraph{How this connects to useful compression.}
Matching every candidate response is unnecessary if the selected result is preserved, and insufficient for accuracy if all matched outputs are wrong. Repeated replacement also requires coverage of reached states and control of error propagation. The same loss-gap identities apply to $\ell_j(a)+\lambda_c\kappa_j(a)$ for specified nonnegative cost $\kappa_j$ and conversion weight $\lambda_c$; a loss bound alone does not bound cost. Operation counts, latency, and training and profiling costs must still be measured separately. Because the candidate and DIRECT use the same fast computation graph, any empirical advantage requires a specified comparison of learning budgets or quality and cost. The reported comparisons of complete training methods establish no such advantage, and the outer selector is fitted separately.
\FloatBarrier
\section{Illustrative Discrete Construction: Protocol and Results}
\label{app:scale-k-r-sealed}

This construction isolates the effect of retained information while fixing the available inputs, summary functions, controller classes, and assigned costs (\cref{fmt:finite-construction}).

\paragraph{Task and allowed actions.}
For $k\in\{2,3,4\}$ blocks, all bits $(A_i,B_i,P_i)$, $i=1,\ldots,k$, are independent and fair. Each continuation query asks for $Y_i=A_i\oplus B_i$ if $P_i=0$, and $Y_i=B_i$ otherwise, where $\oplus$ denotes exclusive OR. The macro reads all $(A_i,B_i)$, never $P_i$, and writes one bit $Z_i$ per block. Query $i$ exposes only $(Z_i,P_i,i)$ to the controller. STOP returns a specified function of these inputs without reading $A_i$, $B_i$, or the label used for scoring; an expert call is allowed to return $Y_i$ and incurs a fee.

Risk is the average error over all queries and uniformly weighted inputs. The total fee per input is $1.60+1.10\,n_{\mathrm{call}}$, where $n_{\mathrm{call}}$ is the number of expert calls for that input and the base fee is charged once. Reported fees are averaged over inputs; dividing by $k$ gives the fee per query.

\paragraph{Fixed summaries and controller classes.}
Arm \textsc{c} retains $Z_i=A_i\oplus B_i$; the generic baseline \textsc{g} retains $Z_i=A_i$, a summary of the same width that is independent of $Y_i$ for either query type. For each arm, one deterministic controller is reused across blocks. Both arms are evaluated over the same class of maps from the four $(Z_i,P_i)$ pairs to $\{\textrm{stop-emit-}0,\textrm{stop-emit-}1,\textrm{call}\}$, giving $3^4=81$ policies. A minimum-fee zero-risk controller for \textsc{c} returns $Z_i$ when $P_i=0$ and calls when $P_i=1$; retaining $Z_i=B_i$ instead would tie its zero-risk fee. Always-expert \textsc{ae} calls at every query. Always-shallow \textsc{as} uses the fixed XOR summary with the best decoder restricted to STOP. Control \textsc{s} retains that same XOR summary but allows a separate choice among the 81 policies for each block; it does not optimize encoders or summaries that depend on the query type.

Exhaustive enumeration covers all $2^{3k}=64,512,4096$ inputs and all policies in the specified classes. Additive risk and fee, with identical block distributions, make separate enumeration for each block exact for \textsc{s}. There is no fitting, sampling, seed, held-out split, or model selection: evaluation covers the entire population. The private evidence package contains the protocol, code, execution records, and aggregate results (\cref{app:reproducibility-details}).

\paragraph{Prespecified criteria and results.}
Before implementation, the protocol fixed three criteria for every $k$. G1 requires \textsc{c} to attain risk $0$ ($\varepsilon=0$). G2 requires \textsc{g}'s minimum risk to remain strictly positive when its expected fee is no greater than \textsc{c}'s minimum zero-risk fee. G3 requires \textsc{c}'s zero-risk fee to be below \textsc{ae}'s, and its risk to be below the minimum risk attainable by \textsc{g} without exceeding \textsc{c}'s fee.

All three hold. For $k=2,3,4$, respectively, \textsc{c}'s minimum zero-risk fees are $2.70/3.25/3.80$, \textsc{g}'s minimum risk under those expected-fee budgets is $0.25$, and \textsc{g}/\textsc{ae}'s minimum zero-risk fees are $3.80/4.90/6.00$. The fee gaps are $1.10/1.65/2.20$, or $0.55$ per query; no tolerance was changed after evaluation. Two stated ties qualify the comparison: \textsc{g}'s zero-risk fee equals \textsc{ae}'s, so G2 and G3 are not independent, and \textsc{s} ties \textsc{c} in minimum zero-risk fee under the fixed-XOR restriction. A lower minimum zero-risk fee for \textsc{s} would contradict the stated enumeration. This tie does not imply that their full risk--fee frontiers coincide.

The result shows a cost advantage from retaining useful information within these finite classes and under the assigned costs. It does not demonstrate learned summaries or measured hardware savings.
\FloatBarrier
\section{Synthetic Methods and Supporting Comparisons}\label{si:synthetic}
\subsection{The Eight-Expert Prototype: Protocol, Controls, and Selector}\label{sec:unified-v3}
\Cref{fmt:prototype,tab:arms} describe the design; \cref{app:synthetic-model,app:synthetic-selection} specify the architecture, objectives, gradients, and selector. This section gives data and model-selection details. \Cref{app:numerical-companion} indexes the complete numerical summaries.

\paragraph{Development and confirmation.}
Development parent seeds 3301 and 3307 use 4,096 training and 1,024 development examples, with 600 pretraining and 600 adaptation updates. Candidate selection compares only the full objectives with uniform and relevance-weighted response matching. Mean development MSEs with hard macro dispatch are $0.016450$ and $0.016220$, respectively, selecting RELEVANT before confirmation. These fits are recorded in the protocol but excluded from the supplementary archive and confirmation means.

Confirmation runs with parent seeds 3407, 3413, 3433, 3449, and 3461 use 8,192 training, 2,048 development, 4,096 IID-test, and 4,096 shifted-test examples drawn from disjoint streams. For $x\sim\mathrm{Unif}([-1,1]^4)$ and $b_*(x)=\sin(2x_1)+x_2/2$, the target is
$y=(\sin(2x_3)+x_1x_2,\ x_3+x_4^2)$ when $b_*(x)\ge0$ and
$y=(\cos(2x_3)-x_1x_2,\ x_3-x_4^2)$ otherwise.
The shifted split multiplies the first two coordinates by $1.25$ before computing $y$; neither the branch identity nor the target is available at deployment. Where applicable, arms share initialization and minibatch streams, and only final checkpoints are evaluated. Fine pretraining and adaptation from the pretrained model each use 1,200 Adam updates, learning rate $0.002$, and batch size 128, with Adam reset between phases; DIRECT uses uninterrupted Adam from raw initialization. All 55 confirmation fits and all choices based on development data were fixed before either test split was evaluated. Uncertainty is measured across parent seeds, not examples or timing repeats. Two-output MSE is recomputed in float64, then averaged over examples and parents as in \cref{eq:contract-aggregation}.

\texttt{REFERENCE} denotes the pretrained fine path, and \texttt{ADAPTED-FINE} denotes continued fine-only training. SHORTCUT-CONT is a local control based on continuation repair, not a reproduction of a published language-model benchmark. The auxiliary unit-span maps are not individually anchored to fine-step targets, so the composition failure in \cref{fmt:evidence} does not identify a unique cause.

\paragraph{Budgets and baseline comparisons.}
The update budgets are 2,400 for DIRECT-DATA and 7,910 for DIRECT-COMPUTE. The latter matches the candidate's forward-affine work for pretraining and adaptation up to a $0.01047\%$ rounding shortfall, excluding backward, nonlinear, optimizer, memory, and dispatch work. Each method is charged for pretraining when evaluated alone; total study cost counts the shared parent fit once. The macro itself trains for only 1,200 updates. A stronger comparison of complete training methods requires candidate mean IID MSE at most $95\%$ of each of JOINT-END, DIRECT-DATA, DIRECT-COMPUTE, SHORTCUT-CONT, and its corresponding ROUTER and FROZEN controls. This practical criterion differs from the quality rule against both fine references and is not a powered superiority test.

\paragraph{Selectors for frozen models.}
A separate protocol, fixed before test evaluation, fits both a primary MLP selector and a mandatory linear control for five frozen variants: RELEVANT, JOINT-END, SHORTCUT-CONT, ROUTER-RELEVANT, and FROZEN-RELEVANT. Across five parent seeds, this gives 50 fits. Each model remains frozen while its selector is trained. \Cref{app:synthetic-selection} specifies available inputs, loss targets and their generation costs, thresholds fixed on development data, and acceptance rules. The offline oracle uses development labels and passes the headroom criterion after charging 96 selector MACs, triggering all selector controls under the protocol. Labels remain unavailable at deployment. All ten policies with thresholds selected on development data (five variants and two selector types) fail the quality rule. Broader schedules, a real second domain, and generative flow matching were not tested.

\paragraph{Verification limits.}
The historical audit recomputed 1,945,600 fixed-path per-example MSEs and 409,600 selector training labels after the fits were fixed. These counts describe software verification, not independent replications or new checks for this revision. Private packages retain protocols, summaries, thresholds, timing, and verification records. Historical prediction arrays and checkpoints are not supplied; hashes cannot replace them. \Cref{fmt:reproducibility} describes availability.
\subsection{Synthetic Model and Training Objectives}\label{app:synthetic-model}
This section specifies the architecture, training surrogates, and gradient paths used in \cref{fmt:prototype}. \Cref{tab:arms} summarizes the purpose of each control.

\subsubsection{Fine prefix, macro, and shared continuation}
An encoder maps $x\in\R^4$ to $z_0\in\R^{16}$. Eight shared residual experts, each with hidden width 48, form a two-step fine prefix. At step $t\in\{0,1\}$, a router reads the current state and normalized phase $t/2$, selects one expert, and updates the state:
\begin{equation}
 a_t=\operatorname*{arg\,max}_e\rho_e(z_t,t),\qquad
 z_{t+1}=E_{a_t}(z_t),\qquad
 z_F=z_2,\quad q_F=\tfrac12(\mathbf e_{a_0}+\mathbf e_{a_1}).
 \label{fmt:prototype-prefix}
\end{equation}
Here $E_e$ is expert $e$ and $\mathbf e_e$ its one-hot identity vector. The notation $\rho(z,t)$ accepts the step count $t$ and normalizes it internally to $t/2$, as in the implementation. A width-24 macro predicts $(z_M,q_M)=C(z_0,0,2)$, with a softmax occupancy head. Its phase and span arguments are step counts, divided by two internally: start $j$ and span $n$ correspond to $s=j/2$ and $\Delta=n/2$. The deployed macro replaces the full span of two fine steps; it never executes those steps to obtain $q_F$. Without an occupancy anchor, $q_M$ is simply a learned simplex-valued input to the continuation router.

Both branches $b\in\{F,M\}$ use the same continuation router, expert bank, and two-output head:
\begin{equation}
 p_b=\operatorname{softmax}\!\left(W[z_b,q_b,1]+\beta\right),\qquad
 a_b=\operatorname*{arg\,max}_e p_{b,e},\qquad
 o_b(a_b)=H\!\left(E_{a_b}(z_b)\right).
 \label{fmt:prototype-continuation}
\end{equation}
$W$ and $\beta$ are router parameters, $H$ is the task readout, and the final input coordinate is the common normalized continuation phase. Only the selected expert runs at deployment. State and occupancy affect the router; the selected expert and head consume the state alone. The predicted occupancy of the skipped interval is therefore distinct from the one expert actually executed after it.

\subsubsection{Objectives and training arms}
Training evaluates all eight continuation actions and minimizes the probability-weighted mean of their \emph{individual} task losses, $\sum_e p_{b,e}\|o_b(e)-y\|_2^2/2$, averaged over the minibatch. It does not minimize the loss of a probability-averaged output. For prefix softmax probabilities $p_t^\rho$, the straight-through weights are $\widetilde p_t=\mathbf e_{\arg\max p_t^\rho}-\operatorname{sg}(p_t^\rho)+p_t^\rho$, where the stop-gradient operation $\operatorname{sg}$ leaves its value unchanged in the forward pass but blocks derivatives in the backward pass. Both the state update and $q_F=(\widetilde p_0+\widetilde p_1)/2$ use these weights: the forward pass makes hard selections, while the backward pass differentiates the soft routing probabilities. Neither this surrogate nor the soft continuation objective gives an exact gradient of hard dispatch.

Load balancing penalizes nonuniform \emph{minibatch-average soft routing probabilities}, not hard dispatch counts. For minibatch size $B$ and router-call set $\mathcal R$,
\begin{equation}
 \bar p_{r,e}=\frac1B\sum_{i=1}^B p_{r,e}(x_i),\qquad
 \mathcal L_{\mathrm{load}}=\frac1{|\mathcal R|}\sum_{r\in\mathcal R}
 \sum_{e=1}^{8}\bar p_{r,e}\log\!\left(8\max\{\bar p_{r,e},10^{-12}\}\right).
 \label{eq:prototype-load}
\end{equation}
This is a numerically floored version of the mean KL divergence from uniform. The router-call set is $\{\rho_0,\rho_1,F,M\}$ for JOINT-END, SIG, RESP-ONLY, SHORTCUT-CONT, RELEVANT and ROUTER-RELEVANT; $\{\rho_0,\rho_1,F\}$ for fine pretraining and continued fine-only; and $\{M\}$ for both DIRECT controls and FROZEN-RELEVANT. Here $\rho_t$ is a prefix call and $F,M$ are the two continuation calls; the macro occupancy head is not a separate load term.
The development-only uniform full-response arm and its ROUTER and FROZEN variants use the same router-call sets as their relevance-weighted counterparts; only the response weighting differs.

The joint base objective sums fine and macro task losses, adds $0.05$ times the mean squared endpoint error across coordinates, and adds $0.01\mathcal L_{\mathrm{load}}$. FROZEN-RELEVANT omits the constant fine task loss and uses only the load term for the macro continuation; DIRECT uses only macro task and load losses. Fine pretraining and continued fine-only training use fine task and load losses. The full objective adds three penalties, each with coefficient $0.1$:
\begin{enumerate}[leftmargin=*,itemsep=3pt]
\item \emph{Occupancy anchoring}: mean squared error across coordinates between $q_M$ and the current fine occupancy $q_F$, with gradients stopped through $q_F$.
\item \emph{Composition}: mean squared errors across state and occupancy coordinates between $C(z_0,0,2)$ and two chained unit-span calls, $C(z_0,0,1)$ followed by $C(\cdot,1,1)$ at the first call's predicted successor; the two predicted occupancies are averaged.
\item \emph{Response matching}: mean squared error across coordinates between corresponding fine and macro continuation outputs, uniformly averaged over experts or weighted by $w_e\propto\exp[-\|o_F(e)-y\|_2^2/(2T)]$, with $T=0.1$. These relevance weights are detached, meaning that gradients do not pass through them.
\end{enumerate}
All penalties that use mean squared error across coordinates also average over minibatch examples; composition adds the state and occupancy MSEs before applying its coefficient. Endpoint, composition, and response penalties differentiate both operands wherever parameters are trainable. Response matching has no direct gradient through continuation-router logits, which learn through task and load terms. Likewise, SHORTCUT-CONT compares the outputs selected by each path without differentiating the argmax indices. In the occupancy anchor, gradients are stopped only through the fine target; relevance weights are also detached. The fine occupancy target changes during joint adaptation and is not a fixed external teacher. ROUTER-RELEVANT freezes encoder, expert, and head parameters but retains derivatives through their computations to the macro and the two routers. FROZEN-RELEVANT trains only the macro, retaining derivatives through the frozen continuation to its inputs. Only the span-2 macro is anchored to a fine endpoint and occupancy. The unit-span maps serve as auxiliaries for the composition objective; they are not calibrated replacements for individual fine steps. In addition, $C$ does not accept an arbitrary incoming route history. These choices affect how the composition failures should be interpreted.
\subsection{Input-Only Selection: Training and Decision Rules}\label{app:synthetic-selection}
The selector is fitted after the complete paths are frozen. \Cref{fmt:utility} reports the test results; the oracle below measures how much saving is possible on development data when labels are available.

\subsubsection{Label-informed development headroom}
For development example $i$, let $L_F(i)$ and $L_M(i)$ be the losses of the complete fine and macro paths, and define $d_i=L_M(i)-L_F(i)$. With exactly $k$ macro choices among $N$ examples, the minimum empirical loss of a label-informed oracle is
\begin{equation}\label{eq:utility-oracle}
 \overline L_F+\frac1N\sum_{j=1}^{k}d_{(j)},\qquad d_{(1)}\le\cdots\le d_{(N)}.
\end{equation}
Selecting the $k$ smallest differences minimizes this sum: exchanging a selected larger difference for an unselected smaller one cannot increase loss. This bound describes the best empirical loss at a fixed number of macro choices when labels are available offline; a deployment selector cannot use those labels. Equal-sized development sets weight parent seeds equally, and the constraint concerns average cost rather than per-input deadlines.
\paragraph{Oracle headroom is not achieved utility.}
After charging the selector's 96 MACs, the label-informed oracle chooses the macro on 10,180/10,240 development examples at MSE $0.007167$ (limit $0.007174$), saving $43.01\%$ of affine work. This is \emph{development oracle headroom}, not the performance of a fitted selector. It does not predict attainable test performance or permit label access at deployment.

\subsubsection{Input-only selection with thresholds fixed on development data}
Each deployment selector sees only the raw input $x$, predicts the two path losses, and selects the macro when its predicted loss difference lies below a threshold fixed on development data. The primary selector is a $4\to16\to2$ tanh MLP (96 affine MACs), with a linear $4\to2$ control (8 MACs). Both train for 600 Adam updates using loss targets obtained by running both frozen paths offline on the training split; the cost of generating these targets is included. Five frozen model variants, five parent seeds, and two selector kinds give 50 fits \emph{in total}, not 50 per kind. For each variant and selector kind, one threshold is chosen on pooled development data to maximize saving subject to the quality tolerance against both fine references. Four additional thresholds at development-data quantiles are fixed for descriptive curves and never retuned on test data (\cref{sec:unified-v3}).

For a fixed selector policy $\pi$, the split loss averages each selected path's two-output MSE over examples pooled across parent seeds. Equal split sizes give equal weight to each parent. Quality requires $L_{\pi,s}\le1.02\min(L_{\mathrm{own},s},L_{\mathrm{cont},s})$ on the scored split; the number of parents that pass individually is descriptive and does not determine the pooled decision. Acceptance of dynamic selection on IID data additionally requires at least $10\%$ mean affine-MAC saving, including selector work, and end-to-end batch-1 median latency (p50) at most $90\%$ of own-fine latency, comparing the means of parent medians.
\subsection{Synthetic Evidence: Fixed Paths and Replication Summaries}\label{app:synthetic-lookup}

\begin{figure}[!htbp]
\centering
\includegraphics[width=\linewidth]{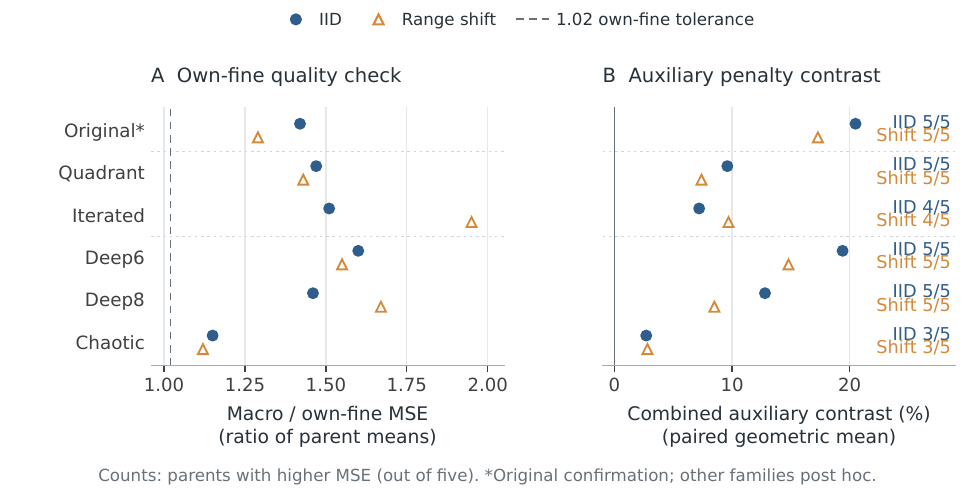}
\caption{\textbf{The quality failure persists; auxiliary effects vary by family.}
Each family uses five parent seeds; IID and range shift are shown separately. (A) Candidate macro MSE divided by own-fine MSE, using ratios of parent means. The $1.02$ line is the own-fine tolerance, a necessary part of the quality rule against both fine references, not the complete acceptance test. (B) Geometric-mean change in macro MSE associated with factor A under the paired contrasts in \cref{app:decomp-shift}; positive values indicate higher error. Factor A combines occupancy anchoring and self-composition and partly reflects the difference in response weighting, so it does not isolate fidelity's causal effect. These diagnostics cover one confirmation family, two replication families, and three families used to search for a stronger fine path. They are not six independent real domains or 60 independent training replications. The numerical supplement reports all parent counts and comparisons with named controls.}
\label{fig:family-quality}
\end{figure}

This section asks whether training a replacement improves on direct training of the same fast architecture, and whether failure to preserve quality persists when the target changes. \Cref{tab:unified-fixed} reports every fixed-path control for the original family and the first two replication families. \Cref{fig:family-quality} compares performance against the quality rule across all six families. \Cref{app:decomp-shift} reports the decomposition diagnostics, \cref{app:factor-decomp} separates the two auxiliary penalties, and \cref{app:teacher-regime} varies the pretraining budget. Reusing parent seeds does not make results from separate fits interchangeable.

\begin{table}[!htbp]
\centering\small
\setlength{\tabcolsep}{3pt}
\caption{\textbf{Direct training of the same fast architecture gives the strongest displayed control.} Complete fixed-path MSE for the original confirmation and first two post-hoc replications, averaged over five parent seeds per family (lower is better). DIRECT-COMPUTE has the lowest mean MSE in each displayed family and split. The candidate's own fine path and continued fine-only model are the quality references; the other controls compare training methods for the fast architecture. Fast paths cost 2,840 forward-affine MACs per example, fine paths 5,176; all three families retain the declared arms and budgets. Shift denotes range shift.}
\label{tab:unified-fixed}
\label{tab:multifamily-fixed}
\begin{tabular}{@{}lrrrrrr@{}}
\toprule
& \multicolumn{2}{c}{Original} & \multicolumn{2}{c}{Quadrant} & \multicolumn{2}{c}{Iterated} \\
\cmidrule(lr){2-3}\cmidrule(lr){4-5}\cmidrule(lr){6-7}
Arm or fine reference & IID & Shift & IID & Shift & IID & Shift \\
\midrule
Pretrained fine & 0.014432 & 0.020641 & 0.093523 & 0.101661 & 0.024047 & 0.037823 \\
Continued fine-only & 0.009712 & 0.012471 & 0.060141 & 0.067100 & 0.011832 & 0.016615 \\
Candidate's own fine & 0.006445 & 0.010635 & 0.066664 & 0.070645 & 0.012356 & 0.018408 \\
\midrule
JOINT-END & 0.007917 & 0.012174 & 0.089037 & 0.093989 & 0.017629 & 0.032582 \\
SIG & 0.009520 & 0.014035 & 0.096632 & 0.099983 & 0.018670 & 0.035291 \\
RESP-ONLY & 0.007721 & 0.011579 & 0.088933 & 0.093510 & 0.017218 & 0.032206 \\
SHORTCUT-CONT & 0.008943 & 0.013466 & 0.094799 & 0.097910 & 0.018796 & 0.033667 \\
RELEVANT (candidate) & 0.009164 & 0.013767 & 0.097732 & 0.100755 & 0.018717 & 0.035837 \\
ROUTER-RELEVANT & 0.020199 & 0.026695 & 0.136945 & 0.146289 & 0.033644 & 0.058613 \\
FROZEN-RELEVANT & 0.024476 & 0.030664 & 0.141470 & 0.151572 & 0.042214 & 0.070139 \\
DIRECT-DATA & 0.005537 & 0.008918 & 0.044928 & 0.052502 & 0.011932 & 0.023200 \\
DIRECT-COMPUTE & 0.003866 & 0.006075 & 0.027293 & 0.035360 & 0.007781 & 0.013154 \\
\bottomrule
\end{tabular}
\end{table}
\FloatBarrier

\paragraph{Does the failure persist across target families?}
The figure uses each candidate's own fine path as the quality reference; the complete evaluation protocol also requires comparison with continued fine-only training. \Cref{fig:family-quality} shows the macro-to-fine MSE ratio for all six families: every family fails even the own-fine part of the $2\%$ rule on both splits. Separately, the candidate loses to both DIRECT controls on every parent--split pair. The factor-A panel shows that the effects of the auxiliary objectives still vary across families.

\FloatBarrier
\subsection{Continuation-Gap Decomposition and Parent-Level Spread}\label{app:decomp-shift}
For frozen paths, the exact identity in \cref{fmt:continuation-decomposition} decomposes macro loss minus its own fine-path loss into fixed-action and reallocation terms. A negative reallocation term offsets replacement error but does not establish superiority to DIRECT. Regret compares each macro with the best action in its own offline oracle menu. DIRECT has no trained fine path to decompose. \Cref{fig:synthetic-diagnostics}(A) shows all parent seeds for the original family's IID split; complete IID and shifted decompositions and oracle diagnostics are reported in the numerical supplement.

The mean reallocation terms do not support a general claim about routing. Removing parent 3413 changes JOINT-END's and SHORTCUT-CONT's IID mean reallocations from negative to positive. ROUTER-RELEVANT's negative total gap accompanies deterioration of its fine-path MSE from $14.43$ to $28.97$ ($\times10^{-3}$). Under shift, SIG's mean reallocation is $-1.04\times10^{-3}$ despite being positive on $4/5$ parents; the mean magnitude and the number of positive values answer different questions. Neither component alone establishes that the quality rule is met. Independent per-example recomputation would require the unavailable historical arrays or retraining.

\paragraph{Paired factorial contrasts.}
Within each parent, JOINT and RESP abbreviate the macro MSEs of JOINT-END and RESP-ONLY; SIG and RELEVANT denote the corresponding methods' macro MSEs. The following paired comparisons average the changes associated with adding occupancy anchoring and composition ($A$), or response matching ($R$). Each difference of log MSEs is the log of a within-parent loss ratio:
\[
\begin{aligned}
A&=\tfrac12[(\log\mathrm{SIG}-\log\mathrm{JOINT})+(\log\mathrm{RELEVANT}-\log\mathrm{RESP})],\\
R&=\tfrac12[(\log\mathrm{RESP}-\log\mathrm{JOINT})+(\log\mathrm{RELEVANT}-\log\mathrm{SIG})].
\end{aligned}
\]
The percentage effects are $100(e^{\bar A}-1)\%$ and $100(e^{\bar R}-1)\%$, with bars denoting parent means. Exponentiating the average log ratios gives the geometric-mean loss ratio across the corresponding comparisons; subtracting one expresses the proportional change. Positive percentages mean higher task error.
RELEVANT changes response weighting as well as adding the two A terms, so A does not isolate occupancy fidelity. The JOINT-END$\to$SIG contrast is easier to interpret because neither arm uses response matching. In the original family, A raises error on all five parents of both splits; R lowers it on all five IID parents but reverses for one shifted parent. These results describe the tested objectives and do not attribute the changes causally to fidelity alone.

\subsection{Sensitivity to the Joint Penalty Weight}\label{app:sensitivity-a}
This follow-up was designed after the original results, with its protocol fixed before the new runs. It varied the common weight of the anchor and composition penalties over $\lambda\in\{0.02,0.05,0.2\}$, using the same five parents, minibatch streams, and test draws. Its \emph{no helpful regime} rule requires SIG$_\lambda$ to have higher IID error than JOINT-END on at least $4/5$ parents at every tested weight; all weights give $5/5$. The historical $\lambda=0.1$ result was not rerun as part of this sweep under the same runtime. Varying both penalties together does not identify which causes the change or exclude helpful untested weights.

The sweep uses PyTorch 2.8.0, whereas confirmation used 2.14.0. A $\lambda=0.1$ fit exactly reproduces parent 3407, but this check covers only one arm and split. SIG discrepancies between runtimes reach $4.2\%$ on IID and $4.8\%$ under shift for parent 3461; the $+2.6\%$ margin for parent 3413 at $\lambda=0.02$ is smaller. Exact effects and runtime-comparison values are reported in the numerical supplement. \Cref{app:factor-decomp} separates the two penalties at one weight.
\subsection{Multi-Family Replication (Declared Post Hoc)}\label{app:multifamily}
Two replication families were specified after confirmation but before their own runs. Only the target generator changed: \emph{quadrant} is a four-region piecewise map with saturating nonlinearities; \emph{iterated} applies $z_{k+1}=\tanh(A_bz_k+c_b)$ twice with branch-dependent $A_b,c_b$. Architecture, objectives, data sizes, optimizer, eleven arms, DIRECT budgets, timing procedure, order of fixing experimental choices, and candidate were unchanged. Replication requires candidate error above DIRECT-DATA and positive factor A on at least $4/5$ IID parents in each family; both families meet this criterion. \Cref{tab:multifamily-fixed} reports all fixed-path results. The macro-to-fine batch-one latency ratio is $0.475$ in both families, but both fail the quality rule. Checks outside the timer confirmed that predictions on the timing inputs match saved full-batch predictions within relative tolerance $2\times10^{-5}$ and absolute tolerance $2\times10^{-6}$, with identical continuation-expert choices.

A second post-hoc protocol, fixed before its runs, specifies \emph{deep6} and \emph{deep8}, with six and eight branch-dependent tanh iterations, respectively, and \emph{chaotic}, with four logistic-map steps whose parameter $r\in[3.6,3.9]$ depends on branch and input. All use the same candidate and hyperparameters. A fine path qualifies as worth replacing only if its own-fine MSE is lower than DIRECT-COMPUTE on at least $4/5$ IID parents. All three families give $0/5$ on IID and at most $1/5$ under shift. These synthetic families share inputs and output dimension; they are not independent real domains.

\paragraph{What does and does not persist.}
\Cref{fig:family-quality,fmt:evidence} summarize the complete paths' failure to preserve quality and the varying effects of the auxiliary penalties. Across 60 parent--split pairs, DIRECT-DATA beats every structured arm; DIRECT-COMPUTE does so on 59, except iterated IID parent 5461 (RESP-ONLY $0.01469$, DIRECT-COMPUTE $0.01489$). The candidate loses to both DIRECT controls throughout, but outperforms ROUTER-RELEVANT and FROZEN-RELEVANT, which restrict adaptation to fewer components. Reallocation is not uniformly harmful: its candidate IID mean is $-6.31\times10^{-3}$ in quadrant ($1/5$ positive), versus $+2.30\times10^{-3}$ in iterated ($4/5$ positive). Each term compares with its own fine action, not DIRECT. R lowers error on only $2/5$ IID parents in either replication; its mean effects are $+0.4\%$ and $-1.4\%$. The original response-matching benefit therefore does not recur consistently.

All decomposition values, parent comparisons, and exploratory groups defined by fine-path quality appear in the numerical supplement. These outcome-defined groups confound family with fine-path quality, so they cannot establish a causal effect of teacher quality. The inverse ordering between fidelity and task error across the original five arms also fails at longer budgets, and the most faithful arms use both anchor and composition penalties. The follow-ups below vary pretraining budgets and separate the two penalties to examine these questions.
\FloatBarrier
\subsection{Effect of Fine-Path Pretraining Budget}\label{app:teacher-regime}
Could insufficient training of the fine model explain the earlier failures? Here \emph{fine pretraining} is the initial task training of the original model with its two-step fine prefix, before adapting it jointly with the replacement. We increase this stage from 1,200 to 4,800, 9,600, or 19,200 updates ($1,4,8,16\times$), keeping adaptation at 1,200 updates after an Adam reset. Other settings and original arms are unchanged.

This is an exploratory follow-up. Conditions and decision rules were fixed after the multi-family results but before these fits; only a reserved-seed timing run preceded declaration. The five confirmation parent seeds and their previously evaluated test sets are reused, so this is not fresh confirmation. Deterministic CPU runs use Linux x86-64, PyTorch 2.14.0, and one thread per process. Budget comparisons use the rerun $1\times$ condition as their baseline.

\paragraph{Budgets and rules.}
DIRECT-DATA receives 2,400/6,000/10,800/20,400 updates; DIRECT-COMPUTE receives 7,910/17,722/30,805/56,970, matching candidate pretraining-plus-adaptation forward-affine MACs. Two controls treat the pretrained teacher as given and charge only for replacement. DIRECT-ADAPT receives 1,200 updates ($26\%$ of candidate adaptation MACs), and DIRECT-ADAPT-COMPUTE receives 4,640. All DIRECT controls start from the initial random weights used before fine pretraining and use checkpoints from one trajectory on a shared minibatch stream; given-teacher controls are identical across budgets. Matching forward-affine work does not match total training FLOPs.

We call the candidate's jointly adapted fine path worth replacing relative to a named DIRECT control when its MSE is lower on at least $4/5$ IID parents. DIRECT-COMPUTE is the primary comparison, DIRECT-DATA the secondary one, and DIRECT-ADAPT the lenient comparison when the pretrained teacher is treated as given. The last comparison qualifies at matched affine cost only if the fine path also qualifies against DIRECT-ADAPT-COMPUTE. Shifted results are reported separately and cannot override IID.

The declared outcomes distinguish no qualifying condition; factor A raising error on $\ge4/5$ parents in every qualifying condition; factor A lowering error on $\ge4/5$ parents in any qualifying condition, which takes precedence; and otherwise an inconclusive result. The manipulation check requires the longer-pretrained reference to have lower IID error than its $1\times$ counterpart on $\ge4/5$ parents. A macro win against a named control does not imply that the quality rule is met.

\paragraph{Results.}
All 260 fits finish on the first attempt with finite outputs; longer-pretraining manipulation checks pass on $5/5$ parents. The fine path qualifies only against DIRECT-ADAPT, already at $1\times$, never against the other three DIRECT controls. A raises error on at least $4/5$ parents at every budget on both splits. The macro beats the lenient control on $\ge4/5$ IID parents at each longer budget, but loses to the other DIRECT controls on all 40 parent--split pairs, loses to JOINT-END and RESP-ONLY on $\ge4/5$ IID parents per budget, and fails the $2\%$ own-fine rule. Longer pretraining therefore yields neither a qualifying fine path at matched affine cost nor quality-preserving replacement here.

\begin{figure}[!htbp]
\centering
\includegraphics[width=\linewidth]{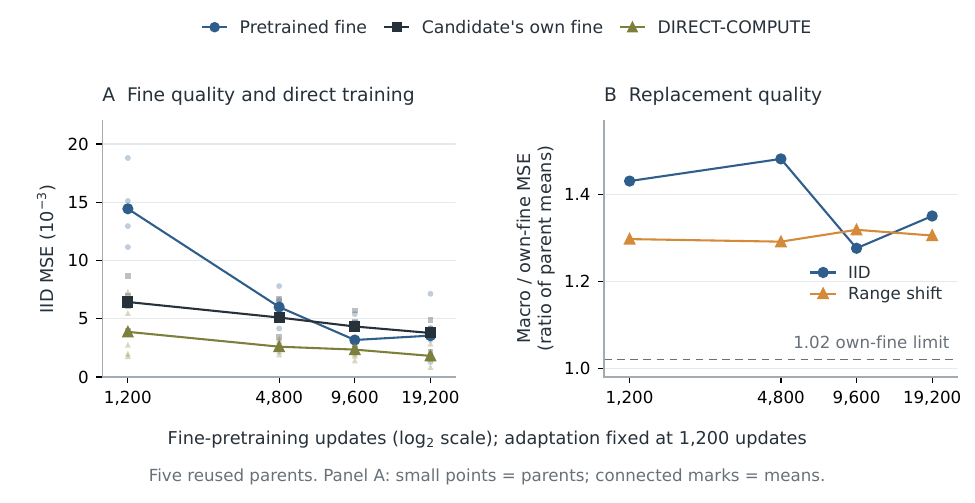}
\caption{\textbf{Longer fine pretraining does not meet the replacement-quality criterion.} (A) IID MSE: small marks show the five reused parents; connected large marks are arithmetic means. DIRECT-COMPUTE matches each budget's pretraining-plus-adaptation forward-affine work. (B) Ratios of macro to own-fine mean MSE on each split. All exceed the $1.02$ own-fine limit, failing a necessary part of the two-reference quality rule. Adaptation remains at 1,200 updates. This exploratory comparison uses the rerun $1\times$ condition, not historical confirmation. Lines connect observations without confidence bands or an interpolation guarantee.}
\label{fig:pretraining-budget}
\end{figure}

\Cref{fig:pretraining-budget} separates improvement of the original computation from preservation of its quality after replacement.

\paragraph{Limits.}
From $8\times$ to $16\times$, pretrained-fine and DIRECT-DATA IID mean errors rise. At both of these budgets, the jointly adapted fine path has higher mean IID error than the pretrained and continued-fine references. Improvements are therefore not monotone. The batch-one macro-to-fine latency ratios are $0.35$--$0.45$, but concurrent CPU jobs make these timings indicative. The prediction and expert-choice checks described in \cref{app:multifamily} pass at every budget. Differences from historical structured-arm results reach $13.4\%$, although the qualification decisions and counts of positive A effects agree. The numerical supplement reports complete counts, effects, means, correlations, and runtime checks. The intervention changes fine-pretraining duration; the experiment does not test other learning-rate schedules, widths, resets, or adaptation budgets.
\subsection{Separating Occupancy Matching from Composition Consistency}\label{app:factor-decomp}
The earlier factor-A comparison added occupancy matching and self-composition together. To distinguish their effects, JOINT-END, ANCHOR-ONLY, COMP-ONLY, and SIG form a $2\times2$ design without response matching. At weight $0.1$, the occupancy anchor $\operatorname{mse}(q_M,q_F)$ uses a fine target with gradients stopped. The composition penalty,
\[
\operatorname{mse}(z_M,z_S)+\operatorname{mse}(q_M,q_S),
\]
compares the span-2 macro with two chained unit-span calls, $C(z_0,0,1)$ then $C(\cdot,1,1)$, also at weight $0.1$. Here $z_S$ is the final state of the chained calls and $q_S$ is their mean predicted occupancy. The penalty measures self-consistency of the macro family rather than fidelity to the fine path. Checks on the reserved seed verified that SIG's objective and gradient equal the base objective plus both penalties. RELEVANT was also fitted and is reported descriptively; it does not enter the declared contrasts or decision rules.

The same five parent seeds receive 1,200 pretraining updates for the primary analysis or 4,800 for the robustness analysis, followed by 1,200 adaptation updates after an Adam reset. Comparisons are paired by parent and minibatch stream; settings and deterministic CPU runtime match \cref{app:teacher-regime}. This exploratory ablation was specified after the original factor-A result and before these follow-up fits. It reuses known parents and previously evaluated test sets, so it is not fresh confirmation. Limited setup and development results were visible, but no decision rule uses them. The reproduction manifest records the exact chronology.

\paragraph{Effects being estimated and decision rules.}
For each penalty separately, $d_p$ is the log of its macro-MSE ratio to JOINT-END within parent $p$. We average these log ratios over the five parents and report $100(e^{\bar d}-1)\%$, the percentage change corresponding to their geometric-mean loss ratio; positive values mean higher error. Counts record the parents with higher error. Let $\ell_{ac}$ be a parent's log MSE, with $a$ indicating whether anchoring is included and $c$ whether composition is included: $\ell_{00}$ is JOINT-END, $\ell_{10}$ is ANCHOR-ONLY, $\ell_{01}$ is COMP-ONLY, and $\ell_{11}$ is SIG. The main effects average each penalty's log-MSE contrast with and without the other penalty:
\[
\begin{aligned}
\text{Anchor main effect}
&=\tfrac12[(\ell_{10}-\ell_{00})+(\ell_{11}-\ell_{01})],\\
\text{Composition main effect}
&=\tfrac12[(\ell_{01}-\ell_{00})+(\ell_{11}-\ell_{10})],\\
\text{Interaction}
&=\tfrac12[(\ell_{11}-\ell_{01})-(\ell_{10}-\ell_{00})].
\end{aligned}
\]
The interaction is defined here as half the difference between the anchor's log-MSE effects with and without composition; it describes how the two penalties combine on this scale. Diagnostics average per-example $\TV(q_M,q_F)$, mean squared endpoint error across coordinates $\overline{(z_M-z_F)^2}$, mean squared composition error across coordinates $\overline{(z_M-z_S)^2}$, and $\TV(q_M,q_S)$.

The criterion for a harmful effect is higher task error than JOINT-END on $\ge4/5$ parents. We distinguish whether the anchor alone, composition alone, or both individual penalties meet this criterion. If neither meets it individually, harm is classified as \emph{only jointly} when SIG meets the criterion and as inconclusive otherwise. IID results at 1,200 updates determine the conclusion; results at 4,800 extend it only if the classification agrees. Shifted results are reported separately and cannot override IID.

Let $f$ count parents for which ANCHOR-ONLY has lower occupancy distance than JOINT-END, and $j$ count those with both lower distance and higher task error relative to JOINT-END. The statement \emph{more faithful and less accurate} requires $j\ge4$; \mbox{$f\ge4,j\le3$} fails that criterion, and $f\le3$ leaves its fidelity premise unmet. Manipulation checks require the anchor to lower occupancy distance and composition to lower composition error on $\ge4/5$ IID parents, each relative to JOINT-END.

\paragraph{Results and uncertainty.}
All 50 adaptation fits and 10 fine-pretraining runs produce finite outputs; both manipulation checks pass on all five parents at both budgets. Composition alone meets the criterion for harm and the anchor does not, on both splits. On IID, composition raises error on $5/5$ parents and the anchor on $2/5$ at both budgets; shifted counts are $4/5$ versus $3/5$ at 1,200 and $5/5$ versus $2/5$ at 4,800. \Cref{fig:synthetic-diagnostics}B shows the primary IID comparison. Descriptive paired $95\%$ $t$-intervals, computed on log ratios and then back-transformed, are $[-11.7\%,+11.8\%]$ and $[+0.2\%,+29.0\%]$ for anchor and composition at 1,200; at 4,800 they are $[-13.5\%,+5.7\%]$ and $[+2.1\%,+29.7\%]$. These IID intervals are not equivalence tests and do not guarantee that the anchor causes no harm.

For the anchor, $f=5,j=2$ on IID at both budgets. Under shift, $f=5$ at both budgets; $j=3$ at 1,200 updates and $j=2$ at 4,800. Despite better fidelity, ANCHOR-ONLY fails the $2\%$ own-fine tolerance: macro-to-own-fine MSE ratios are $1.15/1.12$ (IID/shift) at 1,200 and $1.11/1.17$ at 4,800. Composition accounts for most of SIG's IID log-MSE increase at 1,200 and all at 4,800; this result is specific to these objectives and does not give a general causal account of fidelity.

REFERENCE, JOINT-END, SIG, and RELEVANT match the pretraining-budget study's checkpoints bitwise (40/40); independent per-parent analysis reproduces the recorded statistics. SIG is worse than JOINT-END on $5/5$ IID and $4/5$ shifted parents at 1,200, compared with $5/5$ on both splits historically. The numerical supplement reports complete arm metrics, factorial effects, and conditional contrasts. This study covers one family, weight $0.1$, and two budgets. It does not separate state from occupancy composition, resolve the contrast involving response matching, or separate the penalties at each weight in the joint-weight sweep.
\FloatBarrier
\section{Language-Model Methods and Complete Comparisons}\label{app:realmodel}

Both studies use the residual affine substitute and six arms defined in \cref{fmt:realmodel}. Original weights are frozen, dropout is disabled, and gradients pass through the frozen continuation to the substitute. Sequences are unpadded, use causal attention, and restart positions at zero. For tokens $z_0,\ldots,z_{T-1}$, NLL scores $-\log p(z_t\mid z_{<t})$ for $t=1,\ldots,T-1$; losses are pooled over scored tokens rather than averaged over passages. Boundary MSE divides total squared error by the number of input positions times model width. Output KL uses the full model as its first argument. The studies differ in sampling, KL positions, ridge intercept regularization, and uncertainty estimation, as specified below. All boundaries in a study share one pretrained checkpoint; neither boundaries nor sampling-order seeds are independent pretrained models. Exact versions and extended results are documented in the source archive's reproduction manifest and numerical supplement.

\subsection{GPT-2: sampling, predictions, and provenance}

\paragraph{Fitting and selection.}
Each Adam arm uses 400 updates of four 512-token windows (819,200 sampled positions), 20 warmup updates, learning rate $3\times10^{-4}$, and gradient clipping at 1.0. The learning rate was selected once using CONT-ZERO, not separately for each objective: $10^{-4}$, $3\times10^{-4}$, and $10^{-3}$ were compared after 100 steps at $(6,2)$ on the first eight non-overlapping 1024-token WikiText-2 validation windows. CONT-LS checkpoints are steps $0,25,50,100,200,400$. Objectives, metrics, predictions, and falsification criteria were declared before the runs, and the test procedure was fixed. Seeds 0 and 1 change only minibatch sampling. CONT-ZERO uses both seeds except at $(4,4)$, where seed 0 was not run; \cref{tab:realmodel} specifies availability for the other arms. Matching updates and data does not equalize backward or total training cost.

\paragraph{Data provenance and change of environment.}
WikiText-2 whitespace-only rows are discarded and the remainder joined with two newlines. LAMBADA uses the first 1,000 English test entries. Historical model, tokenizer, and dataset revisions and fingerprints were not retained; current versions cannot recover them. Seed 1 and FID-SGD ran on a second Apple Silicon/MPS machine under a change of environment declared in advance. Reported cross-machine MSE differences were at most $3\times10^{-4}$; the smallest decisive NLL margin was $6\times10^{-3}$. These are different metrics and cannot serve as interchangeable error tolerances.

\paragraph{Ridge fit and sampling.}
For fitting positions $j=1,\ldots,N$, let $x_j=h_l^{(j)}$ and $d_j=h_{l+k}^{(j)}-h_l^{(j)}$. FID-LS minimizes
\[
    \sum_{j=1}^{N}\|Wx_j+b-d_j\|_2^2
    +10^{-3}\bigl(\|W\|_F^2+\|b\|_2^2\bigr).
\]
This is an unnormalized sum of squares with penalized bias; neither activations nor the Gram matrix are divided by sample or coordinate count. With rows $x_j^\top$ in $X$, rows $d_j^\top$ in $D_{\mathrm{fit}}$, and $\widetilde X=[X,\mathbf{1}]$, the coefficient matrix is $(\widetilde X^\top\widetilde X+10^{-3}I_{769})^{-1}\widetilde X^\top D_{\mathrm{fit}}$. The Gram matrix and right-hand side are formed in float32, the system is solved in float64, and parameters are stored in float32. Minimizing this fitting objective need not minimize unregularized held-out MSE.

Ridge uses seed 12345 and 97 batches of four 512-token windows ($N=198{,}656$, the 200{,}000-position cap rounded down to complete batches). Adam uses 400 such batches with the run seed. Both sample window starts uniformly with replacement from $0,\ldots,T_{\mathrm{train}}-514$, where $T_{\mathrm{train}}$ is the flattened training length; windows may overlap. Per window, MSE and KL use all 512 positions and NLL uses 511 next-token targets.

\paragraph{Scoring and aggregation.}
GPT-2 uses no additional selection mask. Unlike Qwen3, its KL includes each sequence's final position; NLL leaves the final logit unused. Positions also restart for each chunk in the exploratory text-likelihood evaluation.

WikiText-2 test uses the first 32 non-overlapping 1024-token windows. Thus the NLL denominator is $32(1024-1)=32{,}736$, the KL denominator is $32\times1024=32{,}768$ (including final positions), and MSE divides by a further 768 hidden coordinates. Training applies the same rule to four 512-token windows: denominators $4\times511$, $4\times512$, and $4\times512\times768$, respectively.

LAMBADA passages are tokenized individually, truncated to their last 1024 tokens if needed, and skipped if shorter than two tokens. Saved evaluations contain 1{,}000 passages and 82{,}641 NLL targets. To score the last word, the text after the final literal space is separately tokenized with one leading space, yielding subword count $L$. A passage is correct iff all $L$ teacher-forced argmax predictions equal its final $L$ token IDs; accuracy divides by evaluated passages. Later subwords therefore receive true earlier subwords, rather than free-running predictions.

Tables average available seeds equally within each arm and boundary. Counts for the declared predictions require every available CONT-ZERO seed to satisfy the stated comparison. P4 correlations pool each arm's final per-seed metrics and CONT-LS steps $25,50,100,200$ within each boundary, without duplicating steps 0 and 400. These measurements assess quality alone and do not test the complete quality--cost criteria.

\paragraph{Declared predictions and outcomes.}
P1 requires FID-LS to have lower boundary MSE but higher NLL than CONT-ZERO on both test distributions at at least four boundaries: it fails, $0/5$ (IID condition $5/5$, shift condition $0/5$). P2 requires every available CONT-ZERO seed to improve IID NLL with MSE at least as large as FID-LS at at least four boundaries: it holds, $5/5$. P3 compares CONT-LS endpoints: step-400 MSE must exceed step-0 MSE while IID NLL falls for every available seed at at least four boundaries. It holds, $5/5$; the six-checkpoint trajectory is archived in the numerical supplement. P4a predicts positive KL--IID-NLL correlation at at least four boundaries: it fails, $0/5$, with Spearman correlations $-0.19$ to $-0.88$. P4b predicts nonpositive boundary-MSE--IID-NLL correlation at at least three: it holds, $5/5$, with correlations $-0.25$ to $-0.59$. The separate falsification criterion, FID-LS IID NLL no higher than CONT-ZERO at at least three boundaries, is not triggered ($0/5$). These pooled associations are descriptive and do not establish an ordering that holds across distributions.

LAMBADA NLL and accuracy remain distinct endpoints: FID-LS beats CONT-ZERO on NLL at all five boundaries, but their accuracy ordering reverses at only two. \Cref{tab:realmodel} reports NLL for every arm on both distributions; all fidelity, accuracy, and trajectory values are in the numerical supplement. These quality comparisons do not test the complete quality--cost criteria.
\begin{table}[!htbp]
\centering\small
\caption{\textbf{The IID preference for continuation fitting reverses under shift.} GPT-2 task loss for all six arms (NLL in nats per scored token; lower is better), with boundaries $(l,k)$ as columns. CONT-ZERO beats FID-LS at every IID boundary, while FID-LS wins every shifted comparison. The other four arms place this contrast alongside the remaining controls. CONT-ZERO, CONT-LS and END-KL average two runs, except CONT-ZERO at $(4,4)$ (seed 1 only); the other arms have one run. The shared, unadapted FULL reference scores $3.444$ (WikiText-2) and $3.611$ (LAMBADA). Additional metrics are in the numerical supplement.}
\label{tab:realmodel}
\begin{tabular}{@{}lrrrrr@{}}
\toprule
\multicolumn{6}{@{}l}{\textbf{A: WikiText-2 (IID)}} \\
Arm & $(6,2)$ & $(4,2)$ & $(8,2)$ & $(4,4)$ & $(6,4)$ \\
\midrule
DROP & 3.879 & 4.015 & 3.914 & 4.910 & 5.010 \\
FID-LS & 3.640 & 3.670 & 3.659 & 4.064 & 4.075 \\
FID-SGD & 3.636 & 3.661 & 3.659 & 4.070 & 4.086 \\
CONT-ZERO & 3.332 & 3.322 & 3.354 & 3.604 & 3.672 \\
CONT-LS & 3.326 & 3.322 & 3.341 & 3.582 & 3.638 \\
END-KL & 3.576 & 3.574 & 3.597 & 3.822 & 3.887 \\
\midrule
\multicolumn{6}{@{}l}{\textbf{B: LAMBADA (shift)}} \\
Arm & $(6,2)$ & $(4,2)$ & $(8,2)$ & $(4,4)$ & $(6,4)$ \\
\midrule
DROP & 3.974 & 3.929 & 3.951 & 4.538 & 4.691 \\
FID-LS & 3.971 & 3.958 & 3.973 & 4.439 & 4.551 \\
FID-SGD & 3.947 & 3.928 & 3.965 & 4.455 & 4.524 \\
CONT-ZERO & 4.137 & 4.054 & 4.281 & 4.454 & 4.667 \\
CONT-LS & 4.202 & 4.071 & 4.410 & 4.511 & 4.803 \\
END-KL & 3.872 & 3.816 & 3.860 & 4.197 & 4.249 \\
\bottomrule
\end{tabular}
\end{table}

\paragraph{Exploratory fitting-text result.}\label{app:realmodel-lookup}
A follow-up at $(6,2)$ changed the fitting text under a protocol declared after the five-boundary results but before its own runs. The update budget and procedure were unchanged. Historical \emph{web}/\emph{mix} records do not identify whether the runs used the requested FineWeb-Edu source \citep{penedo2024fineweb} or the permitted WikiText-103 fallback, so results cannot be attributed to either source. Shifted NLL improves, but no fitted substitute matches or improves upon unadapted FULL on every held-out set: the declared hypothesis of no worse performance fails. The prediction of coming within $0.10$ nats of FULL's LAMBADA NLL misses by $0.005$--$0.012$ (one CONT-LS seed meets it); the deletion-based falsification criterion is not triggered. GSM8K is evaluated by teacher-forced likelihood of question-and-answer text, not problem-solving accuracy \citep{cobbe2021gsm8k}. The source archive's numerical supplement preserves the complete corpus table, sampling and scoring protocol, and trajectory; source provenance remains unresolved.
\FloatBarrier
\subsection{Qwen3: a prespecified within-model comparison}\label{app:qwen3}

\Cref{fmt:realmodel,app:realmodel} give the shared replacement and scoring conventions. This replication compares continuation training with fidelity training at matched updates, then evaluates distribution shift and stronger controls under its own sampling and inference protocol.

\paragraph{What is held fixed.}
Qwen3-1.7B-Base has 28 blocks and width 2,048; the affine substitute has 4,196,352 parameters. Runs use an Apple M5 Pro with a 16-core GPU and 48\,GiB unified memory, Apple MPS, float32, eager attention, and no cache. The reproduction manifest records exact model, tokenizer, and dataset revisions and software versions; the corresponding historical identities for GPT-2 remain unresolved.

DROP and FID-LS run once per boundary; FID-SGD, CONT-ZERO, CONT-LS, and END-KL use sampling-order seeds 1701, 1702, and 1703. Adam uses 256 updates, batch size one, 256-token inputs, learning rate $10^{-4}$, 20 warmup updates, betas $(0.9,0.999)$, epsilon $10^{-8}$, no weight decay, and gradient clipping at 1.0. There is no validation tuning or checkpoint selection: step 128 is diagnostic and step 256 supplies the final scores. Each update stage visits the same 256 windows once (65,536 input positions; 65,280 available next-token positions). FID-SGD supervises hidden-state coordinates at all input positions by MSE.

Here FID-LS minimizes the unnormalized squared error in the boundary increment with penalty $0.001\|W\|_F^2$. Unlike GPT-2, the intercept is \emph{unpenalized}. Sufficient statistics and the solve use CPU float64, with parameters returned in float32. CONT-LS uses the shared ridge fit before its update stage, so matching updates does not match total data exposure or computation.

\paragraph{Which text is compared.}
WikiText-2 supplies its standard training, validation, and IID test splits; the English LAMBADA test set supplies the shifted evaluation. WikiText articles are grouped at level-one headings, joining source rows verbatim and tokenizing without added special tokens. Selection follows source order independently of model outputs. The prepared training pool has 1,024 disjoint 256-token windows from 64 articles; fitting uses its first 256 windows from 21 articles. All articles used for setup checks are excluded from the reported training runs. Text and token hashes detect no exact overlap across splits; this does not rule out semantic overlap or pretraining contamination.

IID and validation retain the first 256 tokens of each eligible article, yielding 61 and 60 prefixes, respectively. The requested quotas of 128 and 64 are not filled by duplication. Shift uses the first 128 LAMBADA passages with at least two tokens, retaining the last 512 tokens when truncation is needed.

\paragraph{What the measurements mean.}
Unlike GPT-2, KL uses the same source positions with valid next-token targets as NLL, leaving the final logit unused. Boundary MSE includes all input positions and 2,048 coordinates. The reproduction manifest gives exact scoring counts. Stochastic entries are arithmetic means across runs with sample SDs; perplexity is exponentiated per run before aggregation. Full-model MSE and KL are set to zero by convention, not measured as reconstruction errors. Tables round means to six significant digits and SDs to three; exact per-run values and validation trajectories remain in the private experimental records.

\paragraph{How the primary comparison is decided.}
For paired IID contrasts,
\[
d_{b,s}=\mathrm{NLL}_{\mathrm{CONT\text{-}ZERO},b,s}-\mathrm{NLL}_{\mathrm{FID\text{-}SGD},b,s},
\]
we average the three boundaries within each seed, then the three seed means. The prespecified criterion requires mean $\le-0.01$, every seed mean $<0$, and negative upper endpoints for both the seed Student-$t$ $95\%$ interval (df 2) and paired-article percentile bootstrap interval. The observed mean is $-0.1901$ nats per scored token; the intervals are $[-0.1976,-0.1826]$ and $[-0.2032,-0.1770]$, respectively. All numerical conditions hold. The bootstrap uses 10,000 draws, seed 909, and common article indices across all nine fitted pairs. The reproduction manifest preserves the seed means, sample SD, and interval endpoints at their original reported precision. All 42/42 replacement runs and the shared FULL reference completed without failures or missing results; original-model parameter hashes were identical before and after execution.

An independent recomputation from 10,707 final per-sample scalar records reproduces the reported aggregate and intervals. This checks arithmetic on saved records; it does not reproduce training. The seeds vary only the order of one fixed pool; boundaries share a model and evaluation set. The article interval is conditional on the fitted substitutes, excludes training-set uncertainty, and is limited by source selection and dependence among articles. Neither interval estimates variation across a population of pretrained models, and they are not combined. Shift and other diagnostics are descriptive, not additional primary tests. All nine shifted contrasts reverse the IID ordering (\cref{fmt:qwen3}).

\paragraph{What the timing can establish.}
Continuation training took longer than fidelity training with the same number of updates at every boundary. Fitting times include diagnostic validation and checkpoint I/O; update intervals also include progress logging and synchronization. CONT-LS is charged for its shared ridge initializer when evaluated alone; summing these costs across seeds would count that fit repeatedly. The reproduction manifest records times by boundary and the exact interruptions.

Execution resumed after host sleep/wake interruptions without a restart or retry. Recorded intervals include these interruptions, and no affected runs are excluded. Memory reports are sampled allocations or process-wide high-water marks, not isolated hardware peaks. Inference uses five measured repetitions after two warmups, with batch size one and no cache, from model input to all output logits. It excludes input preparation, transfer, and generation. These measurements establish neither a speed advantage under stable hardware conditions nor acceleration that preserves quality. The scope is one dense checkpoint, one substitute class, three boundaries, and one budget; historical GPT-2 provenance limitations remain unchanged.

\paragraph{Reading the complete comparison.}
\Cref{tab:qwen3-nll} compares all six arms with the unadapted FULL reference on both distributions, showing the primary IID improvement alongside stronger controls and the reversal under shift.

\begin{table}[!htbp]
\centering\small
\caption{\textbf{The primary IID gain does not identify a universally best fitting objective.} NLL for every Qwen3 arm in nats per scored token; lower is better. CONT-ZERO beats FID-SGD on IID but loses on LAMBADA; FID-LS and CONT-LS provide stronger controls. Every replacement has higher shifted NLL than the shared unadapted FULL reference. Parentheses give sample SD over three sampling-order seeds; entries without parentheses are single deterministic runs. FULL is not an equally adapted control.}
\label{tab:qwen3-nll}
\begin{tabular}{@{}lrr@{}}
\toprule
Arm & WikiText-2 IID & LAMBADA shift \\
\midrule
\multicolumn{3}{@{}l}{\textbf{Boundary $(7,2)$}} \\
DROP & 2.52875 & 3.47792 \\
FID-LS & 2.38001 & 3.37497 \\
FID-SGD & 2.38988 (0.00127) & 3.38201 (0.000867) \\
CONT-ZERO & 2.21467 (0.00331) & 3.54198 (0.00715) \\
CONT-LS & 2.18530 (0.00318) & 3.53587 (0.0198) \\
END-KL & 2.39353 (0.00635) & 3.37113 (0.00732) \\
\midrule
\multicolumn{3}{@{}l}{\textbf{Boundary $(14,2)$}} \\
DROP & 2.44281 & 3.37785 \\
FID-LS & 2.34222 & 3.32403 \\
FID-SGD & 2.36494 (0.000538) & 3.35345 (0.000830) \\
CONT-ZERO & 2.18532 (0.00283) & 3.54874 (0.0230) \\
CONT-LS & 2.15669 (0.00128) & 3.54217 (0.0143) \\
END-KL & 2.35633 (0.00105) & 3.34097 (0.00385) \\
\midrule
\multicolumn{3}{@{}l}{\textbf{Boundary $(21,2)$}} \\
DROP & 2.96416 & 3.64938 \\
FID-LS & 2.62094 & 3.52918 \\
FID-SGD & 2.71419 (0.00169) & 3.64797 (0.00114) \\
CONT-ZERO & 2.49875 (0.00512) & 3.71813 (0.00676) \\
CONT-LS & 2.43803 (0.00502) & 3.77285 (0.00661) \\
END-KL & 2.62999 (0.00314) & 3.51999 (0.00279) \\
\midrule
FULL (shared) & 2.28204 & 3.25889 \\
\bottomrule
\end{tabular}
\end{table}
\FloatBarrier

\begin{figure}[!htbp]
\centering
\includegraphics[width=\linewidth]{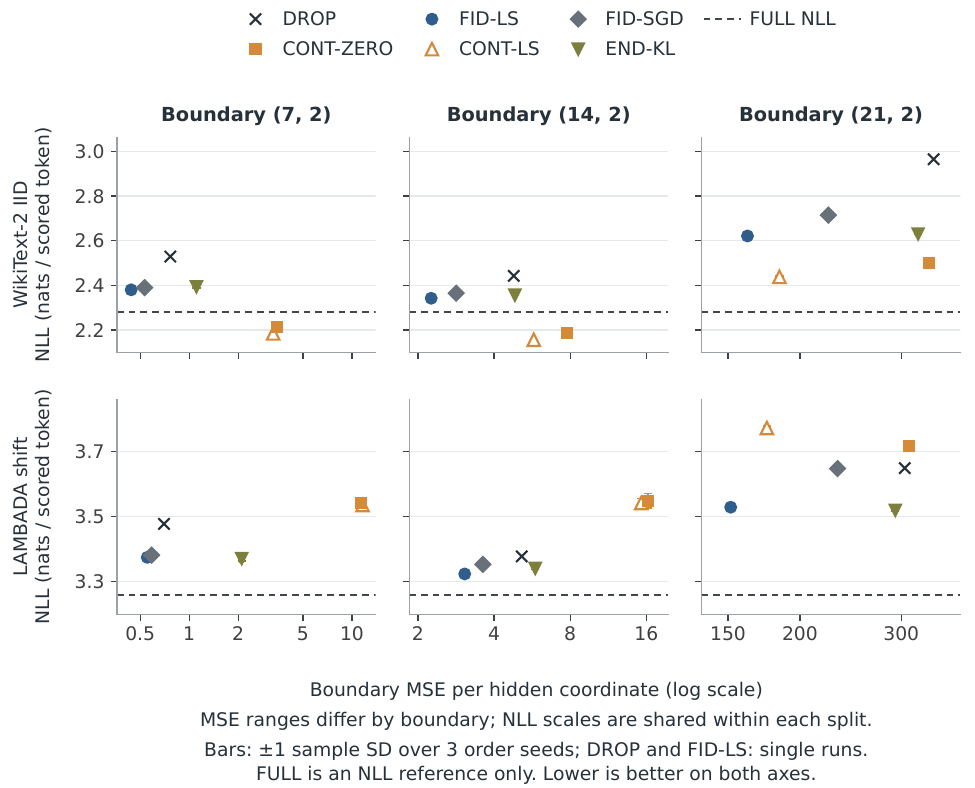}
\caption{\textbf{Fidelity and task quality need separate evaluation.} All six replacement methods appear on both distributions at all three boundaries. Axes and bars are defined in the figure; SD bars may be smaller than markers. FULL is an unadapted NLL reference, not an equally adapted control. Points share one model; they are not independent model replications, and no causal relationship is fitted. Complete MSE, KL, and NLL values are reported in the numerical supplement (\cref{app:numerical-companion}).}
\label{fig:qwen3-fidelity-task}
\end{figure}

\Cref{fig:qwen3-fidelity-task} asks whether lower boundary MSE corresponds to lower NLL, using the same arms, both distributions, and the full-model reference.

\FloatBarrier
\FloatBarrier
\section{Supporting Evidence and Reproducibility}
\label{app:reproducibility-details}

\label{app:numerical-companion}
The source archive's numerical supplement contains complete tables and synthetic follow-up details, values for all GPT-2 arms and the checkpoint trajectory, Qwen3 aggregate metrics, and figure data. Its README defines units, aggregation, and rounding; the manifest records exact revisions and provenance. These summaries do not include per-example evidence. Full experimental packages remain private pending release and licensing. The Qwen3 package retains inputs, per-example metric sums, and substitute checkpoints. Historical arrays and checkpoints and some GPT-2 provenance remain unavailable (\cref{app:realmodel}). Arithmetic checks on saved records are not independent reruns.
\end{document}